\documentclass[lettersize,journal]{IEEEtran}
\usepackage{amsmath,amsfonts}
\usepackage{algorithm}
\usepackage{array}
\usepackage[caption=false,font=normalsize,labelfont=sf,textfont=sf]{subfig}
\usepackage{textcomp}
\usepackage{stfloats}
\usepackage{url}
\usepackage{verbatim}
\usepackage{graphicx}
\usepackage{cite}
\usepackage{graphicx}
\usepackage{booktabs}
\usepackage{multirow}
\usepackage{algorithm}
\usepackage{algcompatible}
\usepackage{algpseudocode}
\usepackage[table]{xcolor} 
\usepackage{booktabs}
\usepackage{amssymb}
\usepackage{bm}

\usepackage{colortbl} 
\usepackage[T1]{fontenc}
\usepackage{textcomp}

\usepackage{float}
\usepackage{verbatim}
\usepackage{graphicx}
\usepackage{textcomp}

\newcommand{\eg}{\textit{e.g.}}
\newcommand{\ie}{\textit{i.e.}}
\begin{document}

\title{Label-Noise Resistant Learning via Optimal Brain Damage Masking}

\author{Xinlei~Zhang,
        Fan~Liu,~\IEEEmembership{Member,~IEEE,}
        Chuanyi~Zhang,~\IEEEmembership{Member,~IEEE,}
        Xiaoying~Ji,
        Wenhui~Wang,
        Wei~Zhou,~\IEEEmembership{Senior Member,~IEEE,}
        and~Yuhui~Zheng,~\IEEEmembership{Member,~IEEE}
        \thanks{ Corresponding author: Fan Liu (fanliu@hhu.edu.cn).}
        
        \thanks{Xinlei Zhang, Fan Liu, Xiaoying Ji, and Wenhui Wang are with the College of Computer Science and Software Engineering, Hohai University, Nanjing, 210098, China.}

        \thanks{Chuanyi Zhang is with the College of Artificial Intelligence and Automation, Hohai University, Changzhou, 213200, China.}

        \thanks{Wei Zhou is with the School of Computer Science and Informatics, Cardiff University, Cardiff CF24 4AG, United Kingdom.}
        
        \thanks{Yuhui Zheng is with the Key Laboratory of Tibetan Information Processing, Ministry of Education, Xining 810008, China.
        }
}

% \markboth{Journal of \LaTeX\ Class Files,~Vol.~X, No.~X}%
% \markboth{IEEE Transactions on Multimedia}
% {Zhang \MakeLowercase{\textit{et al.}}:
% Label-Noise Resistant Learning via Optimal Brain Damage Masking}

\maketitle

\begin{abstract}
Noisy labels are inevitable in real-world multimedia applications. Due to the strong memorization capacity of deep neural networks, these noisy labels cause significant performance degradation. Existing noise-robust methods have mainly focused on robust loss functions and sample selection strategies, with comparatively limited exploration of dynamic architectural adaptation. In this paper, we rethink the role of classifier connectivity under label noise. Intuitively, performance degradation stems from the backpropagation of noisy gradients. Since the final classifier layer acts as the primary gateway for this error propagation, selectively discarding redundant connections can restrict the backpropagation pathways of noisy gradients. Consequently, to identify redundant connections, we leverage the seminal Optimal Brain Damage (OBD) theory from model compression, which posits that parameters causing negligible loss perturbation can be removed. Guided by this principle, we show that masking low-activation edges limits the estimated loss perturbation to preserve the model\textquotesingle s fitting capacity, while tightening a derived upper bound on noise-induced gradient error. To bridge this theoretical insight with practical training, we propose a novel Selective Edge Masking (SEM) mechanism for the fully connected (FC) layer to enhance noise robustness. It can adaptively retain critical edges for information propagation while suppressing gradient errors caused by noisy labels. As a plug-and-play component, SEM can be seamlessly integrated into various noise-robust methods. 
Additionally, we validate SEM\textquotesingle s applicability by applying it to the newly emerged Kolmogorov–Arnold Network (KAN) employed as a classifier. Extensive evaluations on synthetic and real-world benchmarks demonstrate that our approach achieves state-of-the-art performance. 
\end{abstract}


\section{Introduction}
\begin{figure}[!t]
    \includegraphics[width=1.0\linewidth]{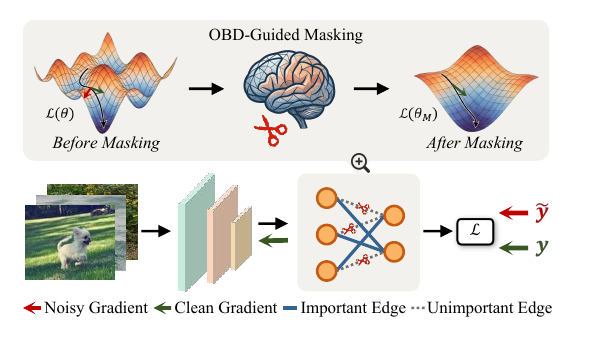}
    \centering    
    % \vspace{-0.5cm}
    \caption{
    Optimal Brain Damage (OBD) theory identifies redundant parameters whose removal induces limited estimated loss perturbation, providing a theoretical basis for preserving fitting capacity in a smaller parameter space $\theta_M$. Guided by OBD, our method adaptively discards redundant connections. It effectively suppresses the backpropagation of noisy gradients induced by corrupted labels $\tilde{y}$ while retaining essential information from clean data $y$. 
    }
    \label{fig1}
    \vspace{-0.1cm}
\end{figure}

Deep neural networks (DNNs) have achieved remarkable performance in various classification tasks~\cite{li2022deep,gong2025randomvig,ma20253d,hong2026unlabeled}. The success largely depends on large-scale, accurately labeled data. However, acquiring high-quality annotations for large-scale multimedia data is prohibitively expensive, inevitably introducing noisy labels into datasets. Extensive studies have shown that training with these corrupted labels can cause significant performance degradation, as DNNs are prone to overfitting on corrupted labels~\cite{sun2021co,zhu2022detecting,sun2022boosting,chen2025towards,wang2026robust}. Consequently, learning with noisy labels has become a critical research focus in deep learning.

Existing noise-robust methods primarily focus on robust loss functions and sample selection~\cite{ghosh2017robust,song2019selfie,wu2023discrimloss,gao2021searching,liu2024learning}. The former achieves risk minimization via loss optimization, which requires careful parameter tuning to balance noise tolerance and sufficient learning for clean data~\cite{wang2021learning,frenay2013classification}. The latter aims to identify clean examples for training, which relies on various heuristic criteria (\eg, small loss~\cite{jiang2018mentornet,shen2019learning}, predicted probability~\cite{yi2019probabilistic,sheng2024foster}).
Additionally, some regularization techniques can also mitigate model overfitting, such as Dropout~\cite{srivastava2014dropout} and DropConnect~\cite{wan2013regularization}. By randomly discarding neurons or connections within the Fully Connected (FC) layer, they implicitly average over an ensemble of subnetworks to reduce overfitting. Nonetheless, their randomness makes it difficult to balance the propagation of noisy and clean signals. 
Distinct from these loss-level or data-level approaches, as illustrated in Fig.~\ref{fig1}, we aim to combat label noise through a simple classifier-level dynamic adaptation strategy, which structurally restricts the noisy gradients while preserving clean information. 

Intuitively, the adverse effects of corrupted labels stem from the backpropagation of noisy gradients. Since these noisy gradients propagate from the final classifier layer toward the upstream feature extractor, the classifier acts as the primary control point for noise propagation. Thus, selectively discarding redundant connections within this layer can structurally suppress the backpropagation of noisy gradients. The critical challenge is precisely localizing these redundant connections without compromising the model\textquotesingle s inherent fitting capacity.

To identify redundant connections, we draw inspiration from Optimal Brain Damage (OBD) theory~\cite{lecun1989optimal}. It posits that parameters causing negligible estimated loss perturbation are redundant and can be removed with limited impact on performance. Although originally from model compression, this principle offers the theoretical basis for discarding redundant connections to mitigate noise-induced overfitting. Guided by OBD, we derive the loss perturbation bound induced by removing specific classifier connections under the standard Cross-Entropy (CE) loss. Our theoretical analysis reveals that this perturbation is upper-bounded by the empirical second moment of the connection\textquotesingle s activation. Thus, edges carrying limited feature information contribute little to the optimization process yet can still transmit noisy gradients during backpropagation. Consequently, dynamically discarding these non-critical edges during training can tighten the gradient-error bound while preserving useful information flow.

To this end, we propose a novel \textbf{Selective Edge Masking (SEM)} mechanism for the widely-adopted FC layer to enhance classifier robustness against noisy labels. Specifically, SEM first dynamically assesses individual connections with the OBD-guided criterion during training. Then, it adaptively masks such low-scoring connections, thereby restricting their gradient backpropagation pathways. This mechanism prioritizes high-scoring critical edges for robust learning, while simultaneously preventing noise propagation through non-essential pathways.
By operating intrinsically within the classifier, our SEM is orthogonal to existing methods that act externally, such as robust loss functions (loss-level) and sample selection strategies (data-level). Consequently, it can be seamlessly integrated with these methods as a plug-and-play module for additional gains. Additionally, to further validate the applicability of our SEM beyond the standard FC classifier, we extend it to the newly proposed Kolmogorov–Arnold Network (KAN)~\cite{liu2024kan} layer serving as the classifier. Experimental results demonstrate that our SEM enhances the robustness of KAN-based models against noisy labels. The main contributions of this paper are summarized as:

\begin{itemize}
    \item 
    To the best of our knowledge, we are the first to explore a novel OBD-inspired masking strategy for noisy label learning. Through theoretical analysis, we demonstrate that masking low-activation connections controls estimated loss perturbation while effectively tightening the upper bound on noise-induced gradient error.
    
    \item 
    From a classifier-level architectural adaptation perspective, we propose a novel Selective Edge Masking (SEM) mechanism tailored for noisy label learning. As a plug-and-play module, it can seamlessly integrate with existing noise-robust methods to achieve better performance.

    \item 
    By extending SEM from conventional FC classifiers to KAN-based classifiers, we provide empirical evidence for the applicability of our structural masking strategy.

    \item 
    We conduct comprehensive evaluations on both synthetic and real-world benchmarks across various noise types. The results demonstrate the effectiveness of our OBD-driven method, which achieves SOTA performance.
    
\end{itemize}

The rest of this article is organized as follows. In Section~\ref{sec2},
we review studies related to our work. In Section~\ref{sec3}, the problem
definition of noisy label learning and the theoretical background of OBD are introduced. In Section~\ref{sec4}, we present the proposed SEM in detail. Section~\ref{sec5} provides comprehensive experimental results and
in-depth analyses of SEM. Finally, the conclusion of this work is drawn in Section~\ref{sec6}.

\section{Related Work}
\label{sec2}
\textbf{Network Pruning.}
It reduces the computational overhead of neural networks by eliminating redundant parameters while preserving performance~\cite{wang2021convolutional, he2023structured}. A fundamental challenge lies in establishing accurate criteria to evaluate parameter importance. To address this challenge, Hessian matrix-based methods are widely recognized for their solid theoretical foundation. The pioneering Optimal Brain Damage (OBD)~\cite{lecun1989optimal} theory links parameter importance to loss perturbation via a second-order Taylor expansion. Subsequent advancements, such as Optimal Brain Surgeon (OBS)~\cite{hassibi1992second}, and Optimal Brain Apoptosis (OBA)~\cite{sun2025optimal}, further refine its numerical precision. While these studies are tailored for extreme network compression, OBD remains the most direct and essential method for modeling how individual parameters perturb the learning objective. Despite its success in model compression, its potential for mitigating label noise remains underexplored.

\textbf{Noisy Label Learning.}
Existing noise-robust methods mainly focus on robust loss functions, sample selection strategies, and regularization techniques. Robust loss functions are designed to achieve noise-tolerant loss. Early study shows that symmetric losses like Mean Absolute Error (MAE) exhibit noise robustness but suffer from slow convergence~\cite{ghosh2017robust}. To address this limitation, methods such as GCE~\cite{zhang2018generalized}, APL~\cite{ma2020normalized}, and ANL~\cite{ye2024active} balance robustness and efficiency through asymmetric transformations and active-passive combinations. Recently, JAL~\cite{wang2025joint} extends this paradigm by introducing a novel asymmetric loss (AMSE) within the APL framework for superior noise tolerance. Alternatively, sample selection aims to filter out noisy samples. Building upon classic multi-network frameworks like Co-teaching~\cite{han2018co}, recent studies introduce more sophisticated filtering criteria. Notably, CA2C~\cite{sheng2025ca2c} introduces a prior-knowledge-free framework for adaptive selection, and Jo-SNC~\cite{sun2025jo} further explores joint consistency evaluation via sample self-consistency and neighbor consistency. Furthermore, regularization techniques enhance model generalization by imposing constraints. Popular methods include Dropout~\cite{srivastava2014dropout} and DropConnect~\cite{wan2013regularization}, which randomly mask hidden units or classifier connections. Building upon these methods, Dynamic DropConnect (DDC)~\cite{yang2025dynamic} mitigates overfitting using gradients to identify critical edges. However, retaining edges with high gradients is unsuitable in noisy settings, as these gradients are largely driven by error signals from noisy labels. To explicitly alleviate noise memorization, advanced regularization methods have been proposed. Among these, CDR~\cite{xia2020robust} identifies critical parameters via gradient-weight products. Despite its effectiveness, evaluating such metrics across the entire network incurs non-trivial computational overhead. Thus, we aim to implement a simple, classifier-level architectural adaptation for robust learning.

\textbf{Kolmogorov-Arnold Networks.} 
Inspired by Kolmogorov-Arnold representation theorem, KAN~\cite{liu2024kan} is a promising alternative to traditional Multi-Layer Perceptron (MLP). Unlike MLP with fixed activation functions at nodes, KAN employs learnable activation functions on edges. While KAN has demonstrated effectiveness in machine learning tasks~\cite{somvanshi2025survey,ta2026fc}, its robustness to noisy labels remains underexplored. 

\section{Preliminaries}
\label{sec3}
\subsection{Problem Definition} 
Consider a $C$-class single-label classification task. Ideally, let $D=\left\{\left({x}_{i}, \mathbf{y}_{i}\right)\right\}_{i=1}^{N}$ denote a clean training set, where $x_i$ is the $i$-th sample, and $\mathbf{y}_{i} \in \{0, 1\}^{C}$ is its true one-hot label. Since acquiring clean data is impractical, we utilize a noisy dataset $D_{\eta}=\left\{\left({x}_{i}, \tilde{\mathbf{y}}_{i}\right)\right\}_{i=1}^{N}$, where $\tilde{\mathbf{y}}_{i}$ may differ from $\mathbf{y}_{i}$. 

A general classification model $f$ is expressed as $f={g} \circ {\psi }$, where visual backbone $\psi$ extracts feature maps for each image $x_i$, and classifier $g$ projects input features to prediction logits over the label space. 
The training objective is to encourage that the global minimizer $f^{*}$ under label noise also serves as the minimizer under clean label supervision~\cite{zhang2018generalized}.

\subsection{Optimal Brain Damage} 
Under the OBD assumptions of local stationarity and a second-order Taylor approximation, removing a parameter $w_q$ (equivalent to imposing a perturbation $\delta w_q = -w_q$) yields the following second-order estimate of the loss perturbation $\delta \mathcal{L}$:
\begin{equation}
\delta \mathcal{L} = \frac{1}{2} {H}_{qq} (\delta w_q)^2 = \frac{1}{2} {H}_{qq} (w_q)^2, {H}_{qq} = \frac{\partial^2 \mathcal{L}}{\partial w_q^2},
\label{eq1}
\end{equation}
where ${H}_{qq}$ denotes the second-order partial derivative with respect to parameter $w_q$. This estimate indicates that a parameter $w_q$ with
negligible estimated loss perturbation ($\delta\mathcal{L}\rightarrow0$) can be regarded as redundant, thereby supporting its removal with limited disruption to the
model\textquotesingle s fitting capacity.

\begin{figure*}[!t]
    \centering
    \includegraphics[width=0.99\linewidth]{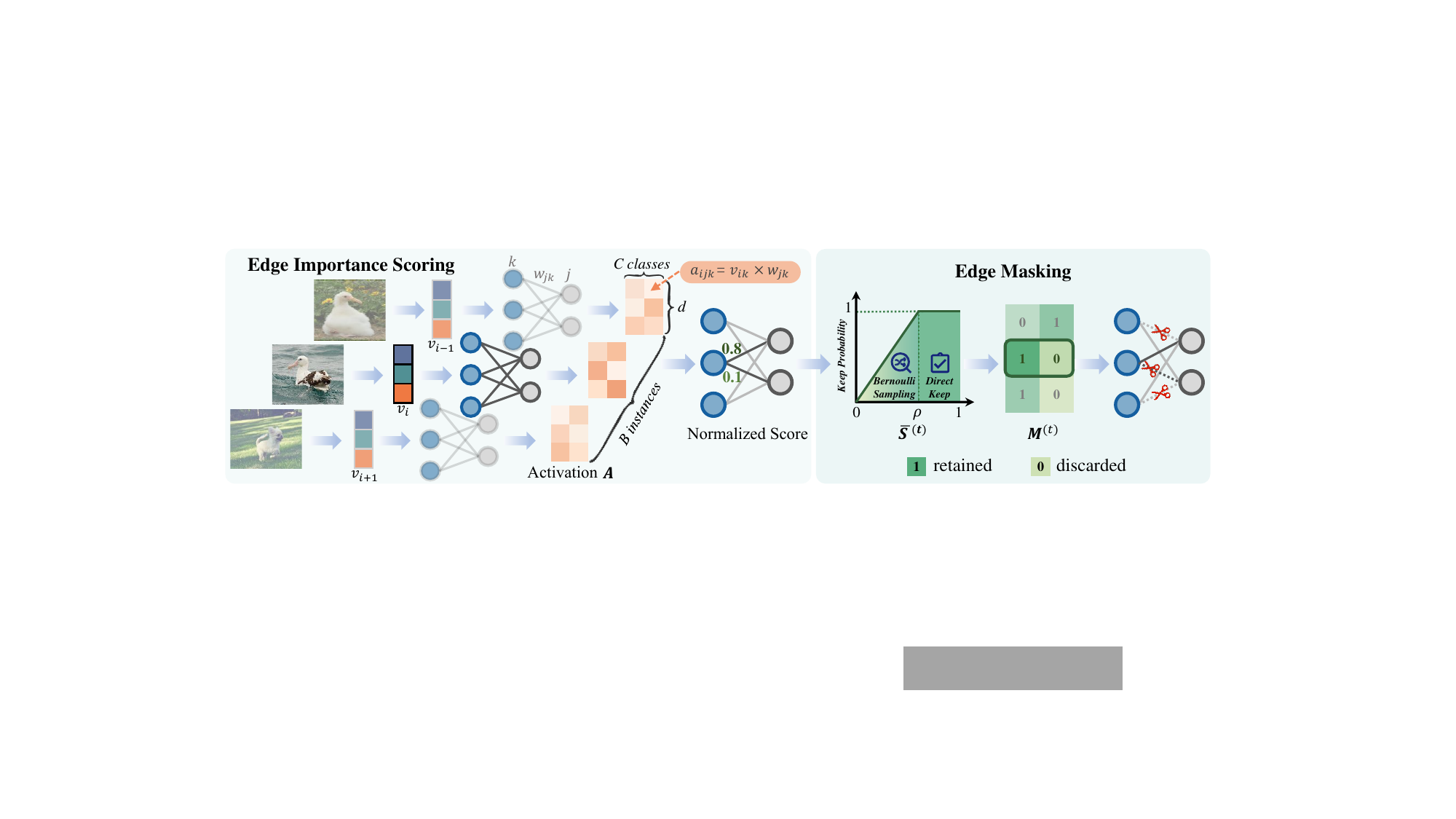}
    \vspace{-0.2cm}
    \caption{
    Overview of our SEM. (i) The edge activation values $\boldsymbol{A}=[a_{ijk}]\in\mathbb{R}^{B\times C\times d}$ are first computed by multiplying the input feature $v_{ik}$ by edge weight $w_{jk}$, where $B$, $C$ and $d$ denote batch size, class count, and feature dimension. Then, the importance of each edge is measured by the Root Mean Square of its activation values along the batch dimension, \ie, $s_{j k}=\sqrt{\frac{1}{B} \sum_{i=1}^{B}a_{i j k}^{2}}$, followed by min–max normalization to obtain Normalized Score $\hat{\textbf{\textit{{S}}}}$. (ii) At each iteration $t$, we adaptively update the edge mask $\textbf{\textit{{M}}}^{(t)}$ via EMA-smoothed score $\bar{\textbf{\textit{S}}}^{(t)}$, which directly determines the keep probability of each edge.}
    \label{fig2}
    \vspace{-0.2cm}
\end{figure*}

\section{Method}
\label{sec4}
In this section, we first present the theoretical motivation for SEM and then describe the proposed method in detail.

\subsection{Theoretical Analysis}
We motivate SEM through two complementary analyses. Proposition 1
upper-bounds the OBD-estimated loss perturbation to identify connections with limited estimated impact on fitting capacity. Building on this criterion, Proposition 2 shows that the resulting mask yields a no-larger derived upper bound on noise-induced gradient error than that of the corresponding dense FC model. Together, they justify OBD-guided selective masking that prioritizes fitting-capacity preservation while controlling noise-induced gradient propagation.

\subsubsection{Fitting Capacity Preservation}
To characterize how masking classifier connections affect fitting
capacity, we leverage OBD to estimate the induced loss perturbation.

Formally, let $\boldsymbol{V}=[v_{ik}]\in\mathbb{R}^{B\times d}$
denote the features extracted by the visual backbone $\psi$ for a
mini-batch of $B$ samples, where $d$ is the feature dimension. Let
$\textbf{\textit{W}}=[w_{jk}]\in\mathbb{R}^{C\times d}$ denote the
classifier weight matrix for $C$ classes. The contribution of
connection $w_{jk}$ to the class-$j$ logit of sample $i$ is
$a_{ijk}=v_{ik}w_{jk}$. Collecting these edge-wise contributions gives the activation tensor
$\boldsymbol{A}=[a_{ijk}]\in\mathbb{R}^{B\times C\times d}$.

\textbf{Proposition 1.} 
Under the standard CE loss and the OBD assumptions, the estimated loss perturbation $\delta\mathcal{L}$ induced by masking a classifier connection $w_{jk}$ is upper-bounded by one eighth of the empirical second moment of its edge activation, \ie,
$\delta\mathcal{L} \leq \frac{1}{8B}\sum_{i=1}^{B}a_{ijk}^{2}$.

\textit{\textbf{Proof:}} 
For a mini-batch of size $B$ and a $C$-class classification task, the CE loss is formulated as
$\mathcal{L}
=
-\frac{1}{B}\sum_{i=1}^{B}\sum_{j=1}^{C}
y_{ij}\log p_{ij}$,
where
$p_{ij}
=
\frac{\exp(z_{ij})}
{\sum_{c=1}^{C}\exp(z_{ic})}$
is the predicted probability of sample $i$ for class $j$, and 
$z_{ij}
=
\sum_{k=1}^{d}v_{ik}w_{jk}+b_j$
is the corresponding classifier logit.

According to Eq.~\ref{eq1}, the loss perturbation induced by removing
connection $w_{jk}$ is estimated as
\begin{equation}
\delta\mathcal{L}
=
\frac{1}{2}
\left(
\frac{1}{B}\sum_{i=1}^{B}
\frac{\partial^2\mathcal{L}_i}{\partial w_{jk}^2}
\right)w_{jk}^2,
\label{eq2}
\end{equation}

where $\mathcal{L}_i$ denotes the CE loss for sample $x_{i}$.  Applying the chain rule (\ie, $\frac{\partial^2 \mathcal{L}_i}{\partial w_{jk}^2} = \frac{\partial^2 \mathcal{L}_i}{\partial z_{ij}^2} \left( \frac{\partial z_{ij}}{\partial w_{jk}} \right)^2$) and substituting $\frac{\partial^2 \mathcal{L}_i}{\partial z_{ij}^2}=p_{ij}(1 - p_{ij})$, Eq.~\ref{eq2} expands to:
\begin{align}
\begin{split}
\delta \mathcal{L}  &= \frac{1}{2} \left( \frac{1}{B} \sum_{i=1}^B \frac{\partial^2 \mathcal{L}_i}{\partial z_{ij}^2} \left( \frac{\partial z_{ij}}{\partial w_{jk}} \right)^2 \right) w_{jk}^2 \\
&= \frac{1}{2B} \sum_{i=1}^B p_{ij}(1 - p_{ij}) v_{ik}^2 w_{jk}^2 \\
&= \frac{1}{2B} \sum_{i=1}^B p_{ij}(1 - p_{ij}) a_{ijk}^2.
\end{split}
\label{eq3}
\end{align}

Since the predicted probability satisfies $p_{ij}\in[0,1]$, it follows that $p_{ij}(1-p_{ij})\leq 1/4$. Let $\Omega_{jk}=\frac{1}{B}\sum_{i=1}^{B}a_{ijk}^2$ denote the empirical second moment of the activation associated with connection $w_{jk}$. The corresponding estimated loss perturbation $\delta \mathcal{L}$ is thus upper-bounded by:
\begin{equation}
\delta \mathcal{L} \leq \frac{1}{8} \left( \frac{1}{B} \sum_{i=1}^B a_{ijk}^2 \right) = \frac{1}{8} \Omega_{jk}.
\end{equation}

Under the diagonal-Hessian approximation, the multi-edge OBD estimate for a realized mask satisfies $\delta\mathcal{L}_{\mathcal{S}}
\leq\frac{1}{8}\sum_{(j,k)\in\mathcal{S}}\Omega_{jk}$,
where $m_{jk}\in\{0,1\}$ is a binary mask variable and
$\mathcal{S}=\{(j,k)\mid m_{jk}=0\}$ denotes the set of masked
connections. This bound motivates SEM to prioritize connections with smaller $\Omega_{jk}$ for masking, thereby limiting the estimated impact on the model\textquotesingle s fitting capacity. The clean confidence results in Fig.~\ref{fig3} (a) provide empirical support for SEM to preserve the model\textquotesingle s fitting capacity.

\subsubsection{Gradient Error Bound}
Following the OBD-guided selection criterion in Proposition 1, we
analyze how the resulting mask affects an upper bound on the
noise-induced gradient error propagated to the visual backbone.

\textbf{Proposition 2.} 
Under the CE loss, given a fixed training state and a realized SEM mask, the derived upper bound on the noise-induced gradient error of the SEM model $f_{\mathrm{SEM}}$ is no larger than that of its unmasked FC counterpart $f_{\mathrm{FC}}$.

\textit{\textbf{Proof:}}
Given the CE loss $\tilde{\mathcal{L}}
=-\frac{1}{B}\sum_{i=1}^{B}\sum_{j=1}^{C}
\tilde{y}_{ij}\log p_{ij}$ under noisy labels $\tilde{y}_{ij}$, its gradient with respect to the visual backbone
parameters $\theta_{\psi}$ is formulated as:
\begin{equation} 
\frac{\partial\tilde{\mathcal{L}}}{\partial\theta_{\psi}}
=
\frac{1}{B}\sum_{i=1}^{B}
\frac{\partial g(v_i)}{\partial\theta_{\psi}}
\left(\mathbf{p}_i-\tilde{\mathbf{y}}_i\right),
\end{equation}
where $\mathbf{p}_i=[p_{ij}]_{j=1}^{C}$ and
$\tilde{\mathbf{y}}_i=[\tilde{y}_{ij}]_{j=1}^{C}$ denote the predicted probability and noisy label vectors in $\mathbb{R}^{C}$, respectively. The feature $v_i$ is extracted by the backbone $\psi$, and $g(v_i)$
denotes the classifier logits. 
We use the column-stacked derivative convention: $
\frac{\partial g(v_i)}{\partial\theta_{\psi}}
=
\left[
\nabla_{\theta_{\psi}}g(v_i)_1,\ldots,
\nabla_{\theta_{\psi}}g(v_i)_C
\right].
$

To quantify the adverse effect of noisy labels, we define the gradient error $\mathcal{D}_f$ as the $\ell_2$-norm deviation between the noise-corrupted gradient
$\frac{\partial\tilde{\mathcal{L}}}{\partial\theta_{\psi}}$ and the clean-label gradient
$\frac{\partial\mathcal{L}}{\partial\theta_{\psi}}$, both computed from the same forward pass:
\begin{equation}
\mathcal{D}_f =\left\|\frac{\partial \tilde{\mathcal{L}}}{\partial \theta_{\psi}}-\frac{\partial \mathcal{L}}{\partial \theta_{\psi}}\right\|_{2}=\left\|\frac{1}{B}\sum_{i=1}^{B} \frac{\partial g\left(v_{i}\right)}{\partial \theta_{\psi}}\left(\mathbf{y}_{i}-\tilde{\mathbf{y}}_{i}\right) \right\|_{2}.
\end{equation}

We analyze gradient backpropagation through the SEM classifier
$g_{\mathrm{SEM}}(v_i)
=(\textbf{\textit{M}}\odot\textbf{\textit{W}})v_i
+\textbf{\textit{b}}$,
where
$\textbf{\textit{M}}=[m_{jk}]\in\{0,1\}^{C\times d}$
is the binary mask matrix,
$\textbf{\textit{W}}\in\mathbb{R}^{C\times d}$
is the learnable weight matrix, and
$\textbf{\textit{b}}\in\mathbb{R}^{C}$
is the bias vector.
Applying the chain rule, the gradient error
$\mathcal{D}_{f_{\mathrm{SEM}}}$ becomes:
\begin{equation}
\begin{aligned}
\mathcal{D}_{f_{\mathrm{SEM}}}
&=
\left\|
\frac{1}{B}\sum_{i=1}^{B}
\frac{\partial g_{\mathrm{SEM}}(v_i)}
{\partial\theta_{\psi}}
(\mathbf{y}_i-\tilde{\mathbf{y}}_i)
\right\|_2
\\
&=
\left\|
\frac{1}{B}\sum_{i=1}^{B}
\frac{\partial v_i}{\partial\theta_{\psi}}
\frac{\partial g_{\mathrm{SEM}}(v_i)}
{\partial v_i}
(\mathbf{y}_i-\tilde{\mathbf{y}}_i)
\right\|_2
\\
&=
\left\|
\frac{1}{B}\sum_{i=1}^{B}
\frac{\partial v_i}{\partial\theta_{\psi}}
(\textbf{\textit{M}}\odot\textbf{\textit{W}})^{\top}
(\mathbf{y}_i-\tilde{\mathbf{y}}_i)
\right\|_2.
\end{aligned}
\end{equation}

To separate the sample-wise gradient errors within the mini-batch, we apply the triangle inequality
($\|\sum_i\mathbf{u}_i\|_2\leq\sum_i\|\mathbf{u}_i\|_2$), which bounds the overall gradient error in terms of the individual error norms:
\begin{equation}
\mathcal{D}_{f_{\mathrm{SEM}}}
\leq
\frac{1}{B}\sum_{i=1}^{B}
\left\|
\frac{\partial v_i}{\partial\theta_{\psi}}
(\textbf{\textit{M}}\odot\textbf{\textit{W}})^{\top}
(\mathbf{y}_i-\tilde{\mathbf{y}}_i)
\right\|_2.
\end{equation}

By applying the matrix-vector norm inequality
(\ie, $\|\mathbf{A}\mathbf{u}\|_2
\leq\|\mathbf{A}\|_F\|\mathbf{u}\|_2$),
we obtain an upper bound on
$\mathcal{D}_{f_{\mathrm{SEM}}}$:
\begin{equation}
\mathcal{D}_{f_{\mathrm{SEM}}}
\leq
\frac{1}{B}\sum_{i=1}^{B}
\left\|\mathbf{y}_i-\tilde{\mathbf{y}}_i\right\|_2
\left\|\textbf{\textit{M}}\odot\textbf{\textit{W}}\right\|_F
\left\|\frac{\partial v_i}{\partial\theta_{\psi}}\right\|_2
\triangleq
\mathcal{U}_{\mathrm{SEM}},
\label{eq9}
\end{equation}
where $\mathcal{U}_{\mathrm{SEM}}$ denotes the derived upper bound on the noise-induced gradient error for SEM. Applying the same derivation to the unmasked FC counterpart yields the bound
$\mathcal{U}_{\mathrm{FC}}$:
\begin{equation}
\mathcal{D}_{f_{\mathrm{FC}}}
\leq
\frac{1}{B}\sum_{i=1}^{B}
\left\|\mathbf{y}_i-\tilde{\mathbf{y}}_i\right\|_2
\left\|\textbf{\textit{W}}\right\|_F
\left\|\frac{\partial v_i}{\partial\theta_{\psi}}\right\|_2
\triangleq
\mathcal{U}_{\mathrm{FC}}.
\label{eq10}
\end{equation}

Since $\textit{\textbf{M}}$ is binary, its masked
weight matrix satisfies:
\begin{equation}
\left\|\textit{\textbf{M}}\odot\textit{\textbf{W}}\right\|_F
\leq
\left\|\textit{\textbf{W}}\right\|_F.
\label{eq11}
\end{equation}

If SEM masks at least one nonzero weight (\ie, $\textbf{\textit{M}}\odot\textbf{\textit{W}} \neq\textbf{\textit{W}}$), the inequality in Eq.~\ref{eq11} is strict. For any training step with a nonzero noise-induced gradient error, the common factor ($\frac{1}{B}\sum_{i=1}^{B}
\|\mathbf{y}_i-\tilde{\mathbf{y}}_i\|_2
\left\|\frac{\partial v_i}{\partial\theta_{\psi}}\right\|_2$)
in Eqs.~\ref{eq9} and~\ref{eq10} is positive. Thus, substituting the strict form of Eq.~\ref{eq11} yields:

\begin{equation}
\mathcal{U}_{\mathrm{SEM}} < \mathcal{U}_{\mathrm{FC}}.
\end{equation}

Accordingly, under the CE loss and for a fixed training state, the realized SEM mask yields a derived gradient-error bound no larger than that of the dense model. Although this bound comparison holds for any binary mask, SEM is distinguished by the OBD-guided criterion in Proposition 1, which prioritizes masking connections with limited estimated impact on fitting capacity. Fig.~\ref{fig3} further provides empirical evidence that SEM preserves clean-data fitting behavior while reducing actual gradient error under the evaluated noise conditions.

\begin{table*}[!t]
\setlength{\tabcolsep}{8.5pt} % Adjust column spacing
\centering
\caption{Comparison with representative robust loss function methods on CIFAR-10 and CIFAR-100 with clean, symmetric (Sym), and asymmetric (Asym) label noise. The results of existing methods are mainly drawn from JAL~\cite{wang2025joint}. Results "mean±std" are reported over 3 random runs, and the best results are highlighted in bold.
Blue-highlighted regions represent the best method.
}
\vspace{-0.1cm}
\begin{tabular}{cc|ccccc|cc}
\toprule
\textbf{Datasets} & \textbf{Methods} & \textbf{Clean} & \textbf{Sym-20\%} & \textbf{Sym-40\%} & \textbf{Sym-60\%} & \textbf{Sym-80\%} & \textbf{Asym-20\%} & \textbf{Asym-40\%} \\ 
\midrule
\multirow{9}{*}{CIFAR-10} 
 & CE & 90.50±0.22 & 75.21±0.39 & 58.05±0.53 & 38.80±0.45 & 19.74±0.40 & 83.05±0.35 & 73.85±0.07 \\
 & GCE~\cite{zhang2018generalized} & 89.66±0.20 & 89.36±0.19 & 82.19±0.84 & 68.01±0.40 & 46.61±0.39 & 85.72±0.22 & 73.36±0.53 \\
 & SCE~\cite{wang2019symmetric} & 91.51±0.24 & 87.65±0.36 & 79.73±0.29 & 61.79±0.72 & 28.01±0.92 & 85.94±0.38 & 74.33±0.56 \\ 
 & NCE+RCE~\cite{ma2020normalized} & 90.80±0.06 & 88.93±0.04 & 85.89±0.31 & 79.89±0.25 & 54.99±2.13 & 88.62±0.29 & 77.94±0.21 \\
 & ANL~\cite{ye2024active} & {91.74±0.18} & 89.68±0.29 & 87.16±0.16 & 81.28±0.63 & 62.28±1.10 & 89.09±0.21 & 77.99±0.40 \\
 & JAL~\cite{wang2025joint} & 91.63±0.21 & 89.95±0.22 & 87.53±0.10 & 82.03±0.18 & {65.43±0.99} & 89.11±0.38 & 79.54±0.34 \\
 \cmidrule{2-9}
 & CE-SEM & 90.28±0.34 & 78.35±0.58 & 69.42±0.70 & 59.55±0.46 & 39.72±0.81 & 85.26±0.55 & 77.50±0.68 \\
 & \cellcolor[HTML]{DAE8FC}\textbf{ANL-SEM} & \cellcolor[HTML]{DAE8FC}{91.70±0.15} & \cellcolor[HTML]{DAE8FC}{90.07±0.34} & \cellcolor[HTML]{DAE8FC}{87.56±0.29} & \cellcolor[HTML]{DAE8FC}{82.21±0.49} & \cellcolor[HTML]{DAE8FC}{64.18±0.72} & \cellcolor[HTML]{DAE8FC}{89.56±0.33} & \cellcolor[HTML]{DAE8FC}\textbf{80.87±0.46} \\
 % \cline{2-8}
 & \cellcolor[HTML]{DAE8FC}\textbf{JAL-SEM} & \cellcolor[HTML]{DAE8FC}\textbf{91.86±0.39} & \cellcolor[HTML]{DAE8FC}\textbf{90.18±0.66} & \cellcolor[HTML]{DAE8FC}\textbf{88.04±0.42} & \cellcolor[HTML]{DAE8FC}\textbf{82.63±0.66} & \cellcolor[HTML]{DAE8FC}\textbf{66.39±1.08} & \cellcolor[HTML]{DAE8FC}\textbf{90.05±0.41} & \cellcolor[HTML]{DAE8FC}{80.11±0.46} \\
 \midrule
\multirow{9}{*}{CIFAR-100} 
 & CE & {70.93±0.77} & 56.47±1.34 & 39.68±0.77 & 22.64±0.53 & 7.82±0.33 & 58.67±0.45  & 41.51±0.12 \\
 & GCE~\cite{zhang2018generalized} & 61.73±1.30 & 60.58±2.51 & 57.35±0.91 & 46.15±1.10 & 20.33±0.31 & 59.19±1.36  & 40.92±0.21 \\
 & SCE~\cite{wang2019symmetric} & 70.57±0.93 & 55.50±0.35 & 40.13±1.48 & 22.23±1.29 & 7.84±0.56 & 57.84±0.57 & 41.58±0.87 \\
 & NCE+RCE~\cite{ma2020normalized} & 68.07±0.70 & 64.57±0.16 & 58.48±0.51 & 46.73±1.00 & 26.94±1.29 & 62.82±0.57 & 41.50±0.39 \\
 & ANL~\cite{ye2024active} & 70.26±0.15 & 66.93±0.09 & 61.58±0.33 & 52.09±0.58 & {28.01±1.06} & 65.96±0.18 & 45.73±0.74 \\
 & JAL~\cite{wang2025joint} & 70.60±0.09 & {68.25±0.39} & {64.11±0.55} & {56.73±0.65} & 22.80±2.11 & {67.90±0.59} & {56.17±0.32} \\
 \cmidrule{2-9}
 & CE-SEM & 70.82±0.36 & 58.71±0.80 & 46.43±0.59 & 36.30±0.67 & 18.63±0.29 & 59.09±0.37 & 42.62±0.16 \\
 & \cellcolor[HTML]{DAE8FC}\textbf{ANL-SEM} & \cellcolor[HTML]{DAE8FC}{70.79±0.30} & \cellcolor[HTML]{DAE8FC}{67.68±0.49} & \cellcolor[HTML]{DAE8FC}{62.63±0.54} & \cellcolor[HTML]{DAE8FC}{52.95±0.77} & \cellcolor[HTML]{DAE8FC}\textbf{29.04±0.97} & \cellcolor[HTML]{DAE8FC}{66.44±0.25} & \cellcolor[HTML]{DAE8FC}{46.07±0.56}\\
 & \cellcolor[HTML]{DAE8FC}\textbf{JAL-SEM} & \cellcolor[HTML]{DAE8FC}\textbf{71.59±0.60} & \cellcolor[HTML]{DAE8FC}\textbf{69.64±0.87} & \cellcolor[HTML]{DAE8FC}\textbf{65.43±0.56} & \cellcolor[HTML]{DAE8FC}\textbf{58.34±1.06} & \cellcolor[HTML]{DAE8FC}{26.14±2.77} & \cellcolor[HTML]{DAE8FC}\textbf{68.93±0.21} & \cellcolor[HTML]{DAE8FC}\textbf{57.52±1.09} \\ 
\bottomrule
\end{tabular}
\label{tab1}
\vspace{-0.2cm}
\end{table*}

\subsection{Selective Edge Masking}
Guided by OBD, we propose Selective Edge Masking (SEM) for dynamic
masking of classifier connections. As illustrated in Fig.~\ref{fig2}, it encompasses two processes: (i) edge importance scoring and (ii) edge masking. Specifically, we first compute a normalized importance score for each edge to quantify its information-carrying capacity. Then, we dynamically mask edges with low importance scores. 

\textbf{Edge Importance Scoring.}
Proposition 1 bounds the OBD-estimated loss perturbation in terms of the empirical second moment
$\Omega_{jk}=\frac{1}{B}\sum_{i=1}^{B}a_{ijk}^{2}$. However, directly using $\Omega_{jk}$ as an edge score produces a highly skewed score distribution because squaring the activations magnifies their dynamic range (see Fig.~\ref{fig4} in the Experiments Section). We thus define the Root Mean Square (RMS) score matrix $\textbf{\textit{S}}=[s_{jk}]\in\mathbb{R}^{C\times d}$ by applying the square root to $\Omega_{jk}$, thereby preserving the ordering of edges while compressing the score range:
\begin{equation}
s_{j k}=\sqrt{\frac{1}{B} \sum_{i=1}^{B}a_{i j k}^{2}}.
\label{eq5}
\end{equation}

Then, to evaluate the relative importance of each edge, the normalized score $\hat{\textbf{\textit{S}}}=[\hat{s}_{jk}]$ is computed as: 
\begin{equation}
\hat{s}_{jk} = \frac{{s}_{jk} - \min(\textbf{\textit{S}})}{\max(\textbf{\textit{S}}) - \min(\textbf{\textit{S}})+ \epsilon},
\label{eq6}
\end{equation}
where $\epsilon=10^{-8}$ ensures numerical stability. 
To reduce mini-batch fluctuations, we smooth the normalized scores using
an Exponential Moving Average (EMA), with
$\bar{\textbf{\textit{S}}}^{(0)}
=\hat{\textbf{\textit{S}}}^{(0)}$.
For $t\geq1$, the smoothed score matrix
$\bar{\textbf{\textit{S}}}^{(t)}
=[\bar{s}_{jk}^{(t)}]$ is updated as:

\begin{equation}
\bar{s}_{jk}^{(t)}
=
\beta\bar{s}_{jk}^{(t-1)}
+(1-\beta)\hat{s}_{jk}^{(t)},
\label{eq7}
\end{equation}
where $\beta$ denotes the momentum coefficient. 
Thus, we establish an edge scoring mechanism for subsequent masking.

\textbf{Edge Masking.} We adaptively update the edge mask matrix to control which edges are retained or temporarily discarded. 

Given the smoothed importance score $\bar{s}_{jk}^{(t)}$ and a predefined retention threshold $\rho\in(0,1)$, edges with
$\bar{s}_{jk}^{(t)}\geq\rho$ are deemed critical and retained deterministically. Since masking all edges below $\rho$ may excessively disrupt classifier connectivity, we apply an
importance-based \textit{Bernoulli} sampling strategy. Thus, lower-scoring edges possess lower retention probabilities. Formally, the binary mask matrix $\textbf{\textit{M}}^{(t)}=[m_{jk}^{(t)}]$ is given by:
\begin{equation}
m_{jk}^{(t)} = 
\begin{cases} 
1, &  \bar{s}_{jk}^{(t)} \ge \rho, \\
Bernoulli(\bar{s}_{jk}^{(t)}), & 0 \le \bar{s}_{jk}^{(t)} < \rho.
\end{cases}   
\label{eq8}
\end{equation}

During training, $\textbf{\textit{{M}}}^{(t)}$ is dynamically updated, and the masked weight matrix $\textbf{\textit{M}}^{(t)}\odot\textbf{\textit{W}}^{(t)}$ is used for optimization, where $\odot$ is element-wise multiplication. This operation allows straightforward
integration into existing methods.

Since our SEM selectively masks redundant edges with low activations, it has minimal impact on the prediction logits. Thus, we disable masking
(\ie, $\textbf{\textit{M}}=\textbf{1}$) and employ the learned dense classifier weight matrix $\textbf{\textit{W}}$ for direct inference. Further analysis of expectation scaling at inference is provided in Table~S4 of the Supplementary Material.

\begin{table*}[!t]
\caption{Comparison with robust loss functions on CIFAR-10 and CIFAR-100 under instance-dependent (IDN) noise. The results are mainly drawn from JAL~\cite{wang2025joint}. Results "mean±std" are reported over 3 random runs, and the best results are highlighted in bold.
Blue-highlighted regions represent the best method.
}
\vspace{-0.1cm}
\centering
\setlength{\tabcolsep}{11pt} % Adjust column spacing
\begin{tabular}{lc|ccc|ccc}
\toprule
\multirow{2}{*}{\textbf{Methods}} & \multirow{2}{*}{\textbf{Publication}} & \multicolumn{3}{c|}{\textbf{CIFAR-10}} & \multicolumn{3}{c}{\textbf{CIFAR-100}} \\
\cmidrule{3-8}
 & & IDN-20\% & IDN-40\% & IDN-60\% & IDN-20\% & IDN-40\% & IDN-60\% \\
\midrule
 % CE & \\
 GCE~\cite{zhang2018generalized} & NeurIPS 2018 & 86.66±0.14 & 79.99±0.23 & 51.90±0.13 & 61.43±2.24 & 57.07±1.04 & 42.40±0.52\\
 SCE~\cite{wang2019symmetric} & ICCV 2019 & 86.65±0.27 & 74.54±0.34 & 49.83±0.40 & 56.32±0.27 & 39.82±1.43 & 23.19±0.87\\
 NCE+RCE~\cite{ma2020normalized} & ICML 2020 & 89.06±0.26 & 85.11±0.28 & 71.27±0.66 & 64.33±0.46 & 57.53±0.84 & 40.36±0.35\\
 ANL~\cite{ye2024active} & NeurIPS 2023 & 89.71±0.35 & 85.74±0.15 & 69.83±0.38 & 66.89±0.53 & 60.88±0.35 & 48.12±0.48\\
 JAL~\cite{wang2025joint} & ICCV 2025 & 90.01±0.12 & 86.46±0.15 & {75.62±0.18} & 67.51±0.29 & {63.24±0.16} & {51.69±0.68}\\
 \midrule
 ANL-SEM & \multirow{2}{*}{Ours}  & {90.46±0.34} & {86.53±0.37} & {70.91±0.60} & {68.13±0.66} & {62.16±0.72} & {49.78±0.89}\\
 \cellcolor[HTML]{DAE8FC}\textbf{JAL-SEM} & & \cellcolor[HTML]{DAE8FC}\textbf{90.53±0.56} & \cellcolor[HTML]{DAE8FC}\textbf{87.34±0.11} & \cellcolor[HTML]{DAE8FC}\textbf{76.68±0.89} & \cellcolor[HTML]{DAE8FC}\textbf{69.17±0.77} & \cellcolor[HTML]{DAE8FC}\textbf{64.95±0.58} & \cellcolor[HTML]{DAE8FC}\textbf{53.08±1.26} \\
\bottomrule
\end{tabular}
\label{tab2}
\vspace{-0.2cm}
\end{table*}

\begin{table*}[!t]
\caption{Comparison with sample selection strategies on CIFAR-100 and CIFAR80N-O datasets under various noise rates. The results of existing methods are mainly drawn from Jo-SNC~\cite{sun2025jo}. The average test accuracy (\%) is reported over the last 10 epochs, and the best results are highlighted in bold.
}
\vspace{-0.1cm}
\centering
\setlength{\tabcolsep}{9.8pt} % Adjust column spacing
\begin{tabular}{lc|ccc|ccc}
\toprule
\multirow{2}{*}{\textbf{Methods}} & \multirow{2}{*}{\textbf{Publication}} & \multicolumn{3}{c|}{\textbf{CIFAR-100}} & \multicolumn{3}{c}{\textbf{CIFAR80N-O}} \\
\cmidrule{3-8}
 & & Sym-20\% & Sym-80\% & Asym-40\% & Sym-20\% & Sym-80\% & Asym-40\% \\
\midrule
Co-teaching~\cite{han2018co} & NeurIPS 2018 & 43.73±0.16 & 15.15±0.46 & 28.35±0.25 & 60.38±0.22 & 16.59±0.27 & 42.42±0.30 \\
JoCoR~\cite{wei2020combating} & CVPR 2020 & 53.01±0.04 & 15.49±0.98 & 32.70±0.35 & 59.99±0.13 & 12.85±0.05 & 39.37±0.16 \\
Jo-SRC~\cite{yao2021jo} & CVPR 2021 & 58.15±0.14 & 23.80±0.05 & 38.52±0.20 & 65.83±0.13 & 29.76±0.09 & 53.03±0.25 \\
UNICON~\cite{karim2022unicon} & CVPR 2022 & 55.10±0.30 & 31.49±0.18 & 49.90±0.30 & 54.50±0.50 & 36.75±0.14 & 51.50±0.50 \\
CA2C~\cite{sheng2025ca2c} & ICCV 2025 & 63.56±0.09  & 33.56±0.07 & 61.16±0.06 & 64.59±0.10 & 31.27±0.14 & 60.19±0.10 \\
Jo-SNC~\cite{sun2025jo} & IEEE TPAMI 2026 & {65.49±0.15} & {43.45±0.16}  & 61.59±0.10 & {67.81±0.15} & {41.10±0.12} & 62.57±0.19 \\
\midrule
\cellcolor[HTML]{DAE8FC}\textbf{CA2C-SEM} & \multirow{2}{*}{Ours} & \cellcolor[HTML]{DAE8FC}65.22±0.14 & \cellcolor[HTML]{DAE8FC}36.03±0.12 & \cellcolor[HTML]{DAE8FC}\textbf{63.84±0.08} & \cellcolor[HTML]{DAE8FC}65.27±0.23 & \cellcolor[HTML]{DAE8FC}34.96±0.29 & \cellcolor[HTML]{DAE8FC}{62.77±0.15} \\
\cellcolor[HTML]{DAE8FC}\textbf{Jo-SNC-SEM} &  & \cellcolor[HTML]{DAE8FC}\textbf{66.02±0.21} & \cellcolor[HTML]{DAE8FC}\textbf{44.26±0.29} & \cellcolor[HTML]{DAE8FC}{62.43±0.13} & \cellcolor[HTML]{DAE8FC}\textbf{68.39±0.24} & \cellcolor[HTML]{DAE8FC}\textbf{42.06±0.27} & \cellcolor[HTML]{DAE8FC}\textbf{63.41±0.25} \\
\bottomrule
\end{tabular}
\label{tab3}
\vspace{-0.2cm}
\end{table*}

\section{Experiments}
\label{sec5}
In this section, we first detail the experimental settings. Then, we compare SEM with various noise-robust training methods on both synthetic and real-world datasets. Furthermore, we conduct a robustness analysis to validate the capability of SEM to preserve the network\textquotesingle s fitting capacity while effectively mitigating the risk of backpropagating noisy gradients. Finally, we systematically investigate the characteristics of our SEM through comprehensive ablation studies.

\subsection{Experiment Setup}
\textbf{Synthetically Corrupted Datasets.}
The CIFAR-10 and CIFAR-100 datasets both contain $50k$ training and $10k$ test images. Open-set CIFAR80N-O dataset is derived from CIFAR-100~\cite{krizhevsky2009learning}, treating its last 20 classes as out-of-distribution samples. These datasets are corrupted with symmetric, asymmetric, and instance-dependent noise at rates $\eta\in(0, 1)$.

\textbf{Real-World Datasets.} 
WebVision-Mini comprises the first 50 classes of WebVision1.0~\cite{li2017webvision}, using the WebVision-Mini and ImageNet ILSVRC12 validation sets for testing. Clothing1M~\cite{xiao2015learning} contains large-scale clothing images across 14 classes, with 1 million training images and $10k$ test images.  

\textbf{Compared Methods.} 
To demonstrate the plug-and-play capability of SEM, we integrate it into different noise-robust methods, including robust loss (JAL~\cite{wang2025joint} and ANL~\cite{ye2024active}), sample selection (Jo-SNC~\cite{sun2025jo} and CA2C~\cite{sheng2025ca2c}), and semi-supervised learning techniques (DivideMix~\cite{li2020dividemix}). We evaluate our SEM-enhanced models against their original counterparts and other representative methods.
We also compare it with several popular regularization methods, including Dropout~\cite{srivastava2014dropout}, DropConnect~\cite{wan2013regularization}, Dynamic DropConnect~\cite{yang2025dynamic} and CDR~\cite{xia2020robust}.

\textbf{Implementation Details.} 
We align the network architectures with the baseline methods. Specifically, for robust losses~\cite{wang2025joint, ye2024active}, we use an 8-layer CNN for CIFAR-10 and ResNet-34 for CIFAR-100. For sample selection~\cite{sun2025jo, sheng2025ca2c}, we adopt a 7-layer CNN for CIFAR-100 and CIFAR80N-O. For real-world datasets, we employ InceptionResNetV2~\cite{szegedy2017inception} for WebVision-Mini and ResNet-50~\cite{he2016deep} for Clothing1M. We adopt the reported optimal settings for all regularization methods under the same training protocol (\ie, a drop rate of 0.5 for Dropout and DropConnect). Our SEM instead adaptively determines its edge retention ratio during training, with $\rho=0.5$ and $\beta=0.9$ fixed across all datasets. All experiments are conducted on the NVIDIA RTX-3090. Further details and our momentum coefficient $\beta$ analysis are provided in the Supplementary Material.

\vspace{-0.05cm}
\subsection{Evaluation on Synthetic Datasets}
\textbf{Integration with Robust Loss Function Methods.} 
We evaluate SEM integrated with CE and representative loss functions (JAL and ANL). As presented in Table~\ref{tab1}, SEM improves the corresponding baselines across symmetric and asymmetric noise, particularly at higher noise rates. For instance, under Sym-80\% on CIFAR-100, JAL-SEM achieves an accuracy of 26.14\%, outperforming the original JAL of 22.80\% by \textbf{3.34\%}. Additionally, SEM-enhanced models yield gains under instance-dependent noise (Table~\ref{tab2}). The results confirm that SEM effectively boosts noise tolerance across various noise types when coupled with robust loss.

\begin{table*}[!t]
\setlength{\tabcolsep}{8.5pt} % Adjust column spacing
\centering
\caption{
Comparison with popular regularization techniques. 
Test accuracy over 3 random runs is reported.
}
\vspace{-0.1cm}
\renewcommand{\arraystretch}{0.99}
\begin{tabular}{lc|ccc|ccc}
\toprule
\multirow{2}{*}{\textbf{Methods}} & \multirow{2}{*}{\textbf{Scoring Criterion}} & \multicolumn{3}{c|}{\textbf{CIFAR-10}} & \multicolumn{3}{c}{\textbf{CIFAR-100}} \\
\cmidrule{3-8}
 & & Sym-80\% & Asym-40\% & IDN-60\% & Sym-80\% & Asym-40\% & IDN-60\%\\
\midrule
CE & baseline & 19.74±0.40 & 73.85±0.07 & 37.97±0.36 & 7.82±0.33 & 41.51±0.12 & 24.49±0.86 \\ 
CE-Dropout & random & 29.96±1.09 & 74.89±0.43 & 49.31±0.67 & 7.29±0.37 & 42.23±0.47 & 23.93±0.23 \\ 
CE-DropConnect & random & 23.09±0.25 & 76.38±0.33 & 43.39±0.82 & 7.34±0.16 & 42.39±0.37 & 22.86±0.93 \\  
CE-Dynamic DropConnect & gradient & 27.96±0.22 & 77.12±0.32 & 47.18±0.20 & 7.78±0.24 & 39.66±0.60 & 19.75±1.36\\
CE-CDR & gradient$\times$weight & 33.73±0.60 & 75.59±0.46 & 51.94±0.36 & 16.04±0.19 & 40.88±0.69 & 27.89±1.25 \\
\cellcolor[HTML]{DAE8FC}\textbf{CE-SEM (ours)} & \cellcolor[HTML]{DAE8FC}\textbf{activation} & \cellcolor[HTML]{DAE8FC}\textbf{39.72±0.81} & \cellcolor[HTML]{DAE8FC}\textbf{77.50±0.68} & 
\cellcolor[HTML]{DAE8FC}\textbf{61.01±0.15} &
\cellcolor[HTML]{DAE8FC}\textbf{18.63±0.29} &
\cellcolor[HTML]{DAE8FC}\textbf{42.62±0.16} & \cellcolor[HTML]{DAE8FC}\textbf{29.09±0.84} \\
\bottomrule
\end{tabular}
\label{tab4}
\vspace{-0.2cm}
\end{table*}

\begin{table}[!t]
\centering
\setlength{\tabcolsep}{5pt}
\renewcommand{\arraystretch}{0.97}
\setlength{\tabcolsep}{3.8pt}
\caption{Comparisons on the WebVision-Mini dataset. Results marked with * are reimplemented using open-source code. For our SEM-enhanced variants, the mean accuracy at the last epoch over 3 random runs is reported.}
\vspace{-0.1cm}
\begin{tabular}{l c c c c c}
\toprule
\multirow{2}{*}{\textbf{Methods}} & \multirow{2}{*}{\textbf{Publication}} & \multicolumn{2}{c}{\textbf{WebVision}} & \multicolumn{2}{c}{\textbf{ILSVRC12}} \\
\cmidrule{3-6} 
&&  Top-1 & Top-5 &  Top-1 & Top-5 \\
\midrule
DivideMix~\cite{li2020dividemix} & ICLR 2020 & 77.32 & 91.64 & 75.2 & 90.84\\
UNICON~\cite{karim2022unicon} & CVPR 2022  & 77.60 & 93.44 & 75.29 & 93.72\\
NCE~\cite{li2022neighborhood} & ECCV 2022 & 79.50 & 93.80 & 76.30 & 94.10 \\
DISC~\cite{li2023disc} & CVPR 2023 & 80.28 & 92.28 & 77.44 & 92.28 \\
CLIPCleaner~\cite{feng2024clipcleaner}  & ACM MM 2024 & 81.56 & 93.26 & 77.8 & 92.08 \\
JAL$^*$~\cite{wang2025joint} & ICCV 2025  & 69.88 & 86.60 & 66.20 & 86.16 \\
Jo-SNC~\cite{sun2025jo} & IEEE TPAMI 2026 & 82.60  & 94.32 & 80.08 & 93.84\\
\midrule
JAL-SEM & \multirow{2}{*}{Ours} & 71.04 & 87.36 & 67.59 & 87.52\\
\cellcolor[HTML]{DAE8FC}\textbf{Jo-SNC-SEM} &  & \cellcolor[HTML]{DAE8FC}\textbf{84.52} & \cellcolor[HTML]{DAE8FC}\textbf{94.68} & \cellcolor[HTML]{DAE8FC}\textbf{82.12} & \cellcolor[HTML]{DAE8FC}\textbf{94.40}\\
\bottomrule
\end{tabular}
\label{tab5}
\vspace{-0.4cm}
\end{table}

\textbf{Integration with Sample Selection Strategies.}
We integrate SEM with sample selection methods (CA2C and Jo-SNC) and evaluate them on both closed-set and open-set benchmarks. As shown in Table~\ref{tab3}, our models outperform their baseline counterparts across the evaluated settings. For example, CA2C-SEM achieves $36.03\%$ accuracy on CIFAR-100 with Sym-80\%, surpassing the CA2C method by \textbf{2.47\%}. 

Notably, our SEM-enhanced models achieve SOTA performance on both closed-set and open-set datasets.

\textbf{Comparison with Regularization Techniques.}
To further validate the robustness of SEM, we compare it against several representative regularization techniques, including Dropout, DropConnect, Dynamic DropConnect, and CDR. Distinct from these methods, SEM performs classifier-level architectural adaptation through activation-guided dynamic edge masking. Unlike Dropout and DropConnect, which apply fixed masking ratios, SEM adaptively determines edge
retention probabilities from forward activations. It also differs from Dynamic DropConnect, which relies on backward gradients, and CDR, which uses gradient-weight products. We reproduce all baseline methods under the same training protocol with their reported optimal settings. As shown in Table~\ref{tab4}, SEM achieves the best accuracy across the evaluated settings, demonstrating the effectiveness of its classifier-level adaptive masking mechanism.

\vspace{-0.2cm}
\subsection{Evaluation on Real-world Datasets}
We evaluate SEM on real-world noisy multimedia datasets, including WebVision-Mini and Clothing1M, as detailed in Tables~\ref{tab5} and \ref{tab6}, respectively. Overall, the SEM-enhanced models outperform their respective methods. For instance, our Jo-SNC-SEM model yields Top-1 accuracy gains of $\textbf{1.92\%}$ and $\textbf{2.04\%}$ on the WebVision and ImageNet ILSVRC12 validation sets, respectively. Furthermore, Jo-SNC-SEM and DivideMix-SEM achieve the best performance on WebVision-Mini and Clothing1M, respectively.

Comprehensive evaluations on both synthetic and real-world datasets demonstrate that our SEM offers plug-and-play compatibility with existing methods.

\subsection{Robustness Analysis}
We conduct robustness analysis for SEM via \textit{oracle analyses}, where clean labels are utilized strictly as a diagnostic analysis, rather than for training.

\begin{figure}[!t]
    \centering
    \includegraphics[width=1.00\linewidth]{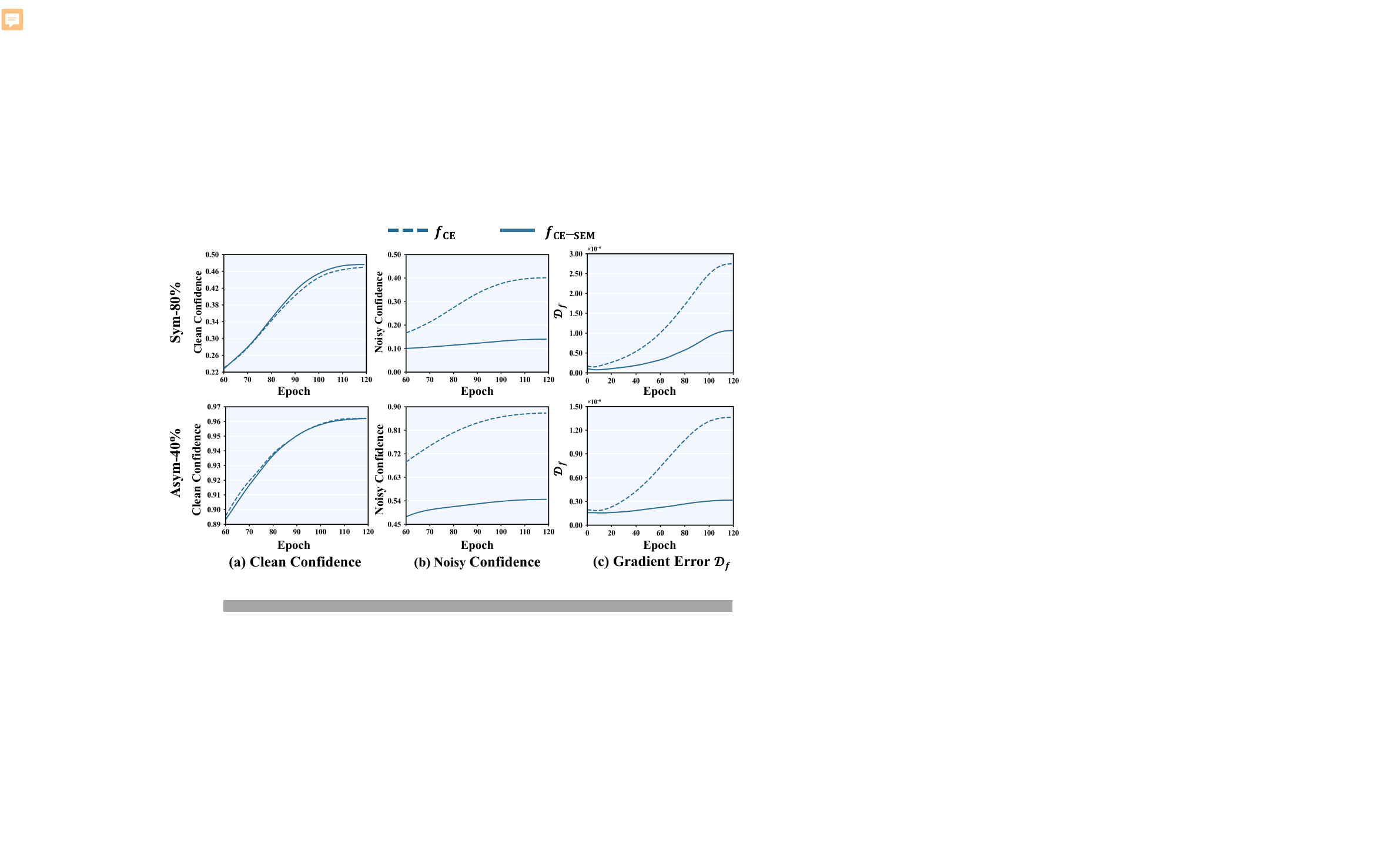}
    \vspace{-0.6cm}
    \caption{
    Robustness analysis on CIFAR-10 under Sym-80\% (top) and Asym-40\% (bottom). (a) Clean Confidence, (b) Noisy Confidence, and (c) Gradient Error $\mathcal{D}_f$. For clear comparison, mid-to-final training results are presented.
    }
    \label{fig3}
    \vspace{-0.3cm}
\end{figure}

\textbf{Confidence Analysis.} 
We conduct a confidence analysis to evaluate the model\textquotesingle s fitting behavior on clean and noisy data during training. Specifically, we define the average prediction probabilities on clean and noisy labels as their respective confidences. Intuitively, a persistently high clean confidence indicates the model\textquotesingle s strong predictive capability on true samples, demonstrating that it can better fit clean data. Similarly, a low noisy confidence reflects the model\textquotesingle s resistance to memorizing corrupted labels, indicating improved noise robustness. As illustrated in Fig.~\ref{fig3} (a), $f_{\text{CE-SEM}}$ maintains clean confidences comparable to its fully connected counterpart $f_{\text{CE}}$ under various noise levels on CIFAR-10. Conversely, $f_{\text{CE-SEM}}$ exhibits substantially lower noisy confidence (Fig.~\ref{fig3} (b)). 

This phenomenon demonstrates that our OBD-guided masking mechanism effectively mitigates overfitting to noisy data while preserving the network\textquotesingle s capacity to fit clean samples. 

\textbf{Gradient Error Analysis}. 
We provide a gradient error analysis to evaluate whether SEM reduces the gradient deviation induced by noisy labels. As illustrated in Fig.~\ref{fig3} (c), we monitor the average $L_2$-norm gradient error $\mathcal{D}_f$ for the parameters of the final backbone block (\ie, block3) throughout training. Specifically, the model performs the backpropagation using both clean and noisy labels to compute respective gradients $\frac{\partial {\mathcal{L}}}{\partial \theta_{\psi}}$ and $\frac{\partial \tilde{\mathcal{L}}}{\partial \theta_{\psi}}$ for recording. 
Only the noisy gradients $\frac{\partial \tilde{\mathcal{L}}}{\partial \theta_{\psi}}$ are utilized for parameter updating. As training iterations increase, our SEM-enhanced models ($f_\text{CE-SEM}$ and $f_\text{JAL-SEM}$) consistently yield lower gradient errors under various noise levels. This empirical evidence demonstrates that our approach effectively reduces the measured gradient magnitude deviation $\mathcal{D}_f$, thereby mitigating the negative effects of noisy labels.

\begin{table}[t]
\centering
\caption{Comparisons on the Clothing1M dataset. Results for competing methods are directly taken from their original papers. For our SEM-enhanced variants, the mean accuracy at the last epoch over 3 random runs is reported.}
\vspace{-0.1cm}
\setlength{\tabcolsep}{13pt} % Adjust column spacing
\renewcommand{\arraystretch}{1.02}
\begin{tabular}{l c c}
\toprule
Method & Publication & Accuracy (\%) \\
\midrule
DivideMix~\cite{li2020dividemix} & ICLR 2020 & 74.76 \\
SOP~\cite{liu2022robust}    & ICML 2022    & 73.5 \\
SSR~\cite{Feng_2022_BMVC}  & BMVC 2022  & 74.83 \\
ELR+~\cite{liu2020early}   & NeurIPS 2020      & 74.81 \\
ProMix~\cite{ijcai2023p494}  & IJCAI 2023      & 74.94 \\
DISC~\cite{li2023disc}  & CVPR 2023 &  73.72 \\
CLIPCleaner~\cite{feng2024clipcleaner}   & ACM MM 2024      & 74.87 \\
JAL~\cite{wang2025joint}         & ICCV 2025  & 70.31 \\
\midrule
JAL-SEM      &  \multirow{2}{*}{Ours} & 71.66 \\
\cellcolor[HTML]{DAE8FC}\textbf{DivideMix-SEM}    &  & \cellcolor[HTML]{DAE8FC}\textbf{75.48} \\
\bottomrule
\end{tabular}
\label{tab6}
\vspace{-0.2cm}
\end{table}

% \vspace{-0.2cm}
\subsection{Ablation Study}
\begin{figure}[!t]
    \centering
    \includegraphics[width=1.0\linewidth]{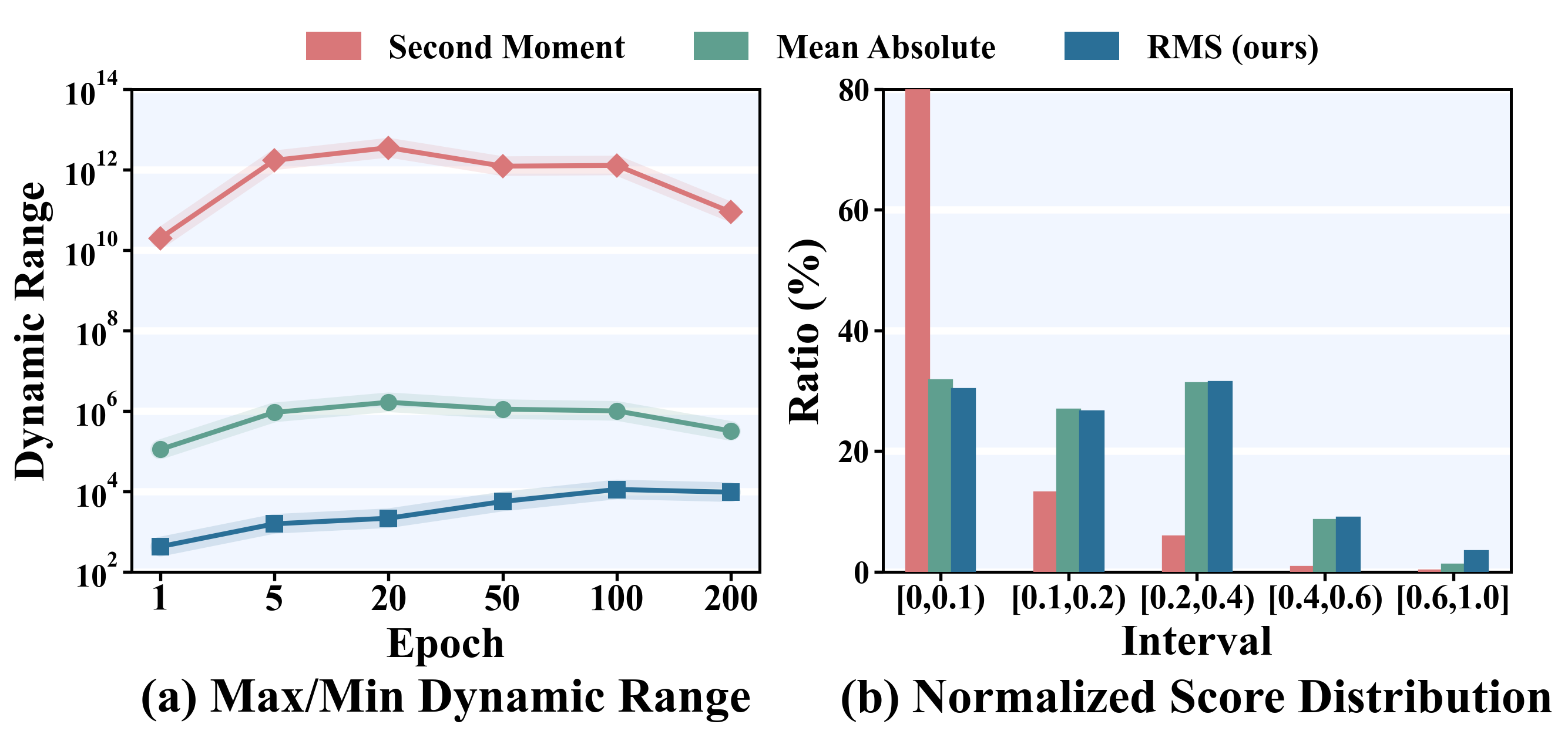}
    \vspace{-0.6cm}
    \caption{ 
    Scoring criteria analysis for CE-SEM on CIFAR-100. (a) Max/Min Dynamic Range and (b) Normalized Score Distribution. 
    } 
    \label{fig4}
    \vspace{-0.2cm}
\end{figure}

\textbf{Scoring Criteria Analysis.} 
To validate the efficacy of the edge importance scoring criteria of our SEM, we evaluate different scoring metrics under 80\% symmetric noise on both CIFAR-10 and CIFAR-100 datasets. As shown in Table~\ref{tab7}, we compare the theoretically derived Second Moment ($\Omega_{jk}$, $\frac{1}{B} \sum_{i=1}^B a_{ijk}^2$), Mean Absolute ($\frac{1}{B} \sum_{i=1}^B |a_{ijk}|$), and Root Mean Square (RMS, $\sqrt{\frac{1}{B} \sum_{i=1}^B a_{ijk}^2}$). Although Proposition 1 establishes the Second Moment as the theoretical OBD bound, directly employing it in practice yields suboptimal performance (\eg, only $7.94\%$ on CIFAR-100). This degradation occurs because squaring the activations magnifies their dynamic range, which induces severe distributional skewness during training. As evidenced in Fig.~\ref{fig4} (a), $\frac{\max(\Omega)}{\min(\Omega)}$ reaches $8.79\times10^{10}$ at the last epoch, showing a dynamic range several orders of magnitude larger than those of Mean Absolute ($3.18\times10^5$) and RMS ($9.56\times10^3$). Consequently, after min-max normalization, this extreme scale makes the scores overly concentrated near zero (Fig.~\ref{fig4} (b)).  While both Mean Absolute and RMS alleviate this issue, Mean Absolute ($\mathbb{E}[\vert{}a\vert{}]$) is generally not a monotonic transformation of $\Omega = \mathbb{E}[a^2]$ across mini-batch samples and may therefore alter the $\Omega$ ordering. In contrast, RMS ($\sqrt{\Omega}$) rigorously preserves the exact ordering induced by $\Omega$, which aligns with its superior performance. Thus, we select RMS as the edge importance scoring criterion to successfully balance the OBD-guided theoretical bound with practical training stability.

\textbf{Masking Component Analysis. } 
We evaluate SEM components under Sym-80\% on CIFAR-10 and CIFAR-100. As demonstrated in Table~\ref{tab8}, solely performing hard-threshold masking to retain critical edges (w/o \textit{Bernoulli}) yields marginal benefits. This result suggests that a rigid hard-threshold masking may excessively discard near-threshold edges, despite their potential benefit to prediction. To mitigate this aggressive masking, we implement \textit{Bernoulli} (0.5), which assigns a fixed retention probability of $0.5$ to all less critical edges. However, this static stochasticity may inadvertently mask vital connections, leading to limited performance gains. Conversely, our score-aware masking strategy (\textit{Bernoulli} ($\bar{\textbf{\textit{{S}}}}$)) enables the adaptive preservation of essential pathways guided by importance scores, achieving substantially superior performance. Furthermore, EMA effectively mitigates score fluctuations across mini-batches, ensuring more stable mask updates and enhanced robustness. 

\begin{table}[!t]
\setlength{\tabcolsep}{9pt} % Adjust column spacing
\renewcommand{\arraystretch}{1.15}
\centering
\caption{
Comparison of various scoring criteria under Sym-80\%, including Second Moment ($\frac{1}{B} \sum_{i=1}^B a_{ijk}^2$), Mean Absolute ($\frac{1}{B} \sum_{i=1}^B |a_{ijk}|$), and Root Mean Square ($\sqrt{\frac{1}{B} \sum_{i=1}^{B}a_{i j k}^{2}}$).
}
\vspace{-0.1cm}
\begin{tabular}{lccc}
\toprule
Method & Scoring Criterion & CIFAR-10 & CIFAR-100 \\ 
\midrule
CE & \multicolumn{1}{c}{baseline} & 19.74±0.40 & 7.82±0.33 \\ 
% \midrule
\multirow{3}{*}{CE-SEM}
 & Second Moment &  39.16±0.69 & 7.94±0.86 \\ 
 & Mean Absolute &  39.49±0.75 & 14.49±1.47 \\  
 & \cellcolor[HTML]{DAE8FC}\textbf{Root Mean Square} & \cellcolor[HTML]{DAE8FC}\textbf{39.72±0.81} & \cellcolor[HTML]{DAE8FC}\textbf{18.63±0.29}\\
\bottomrule
\end{tabular}
\label{tab7}
\vspace{-0.4cm}
\end{table}

\begin{figure}[!t]
    \centering
    \includegraphics[width=1.0\linewidth]{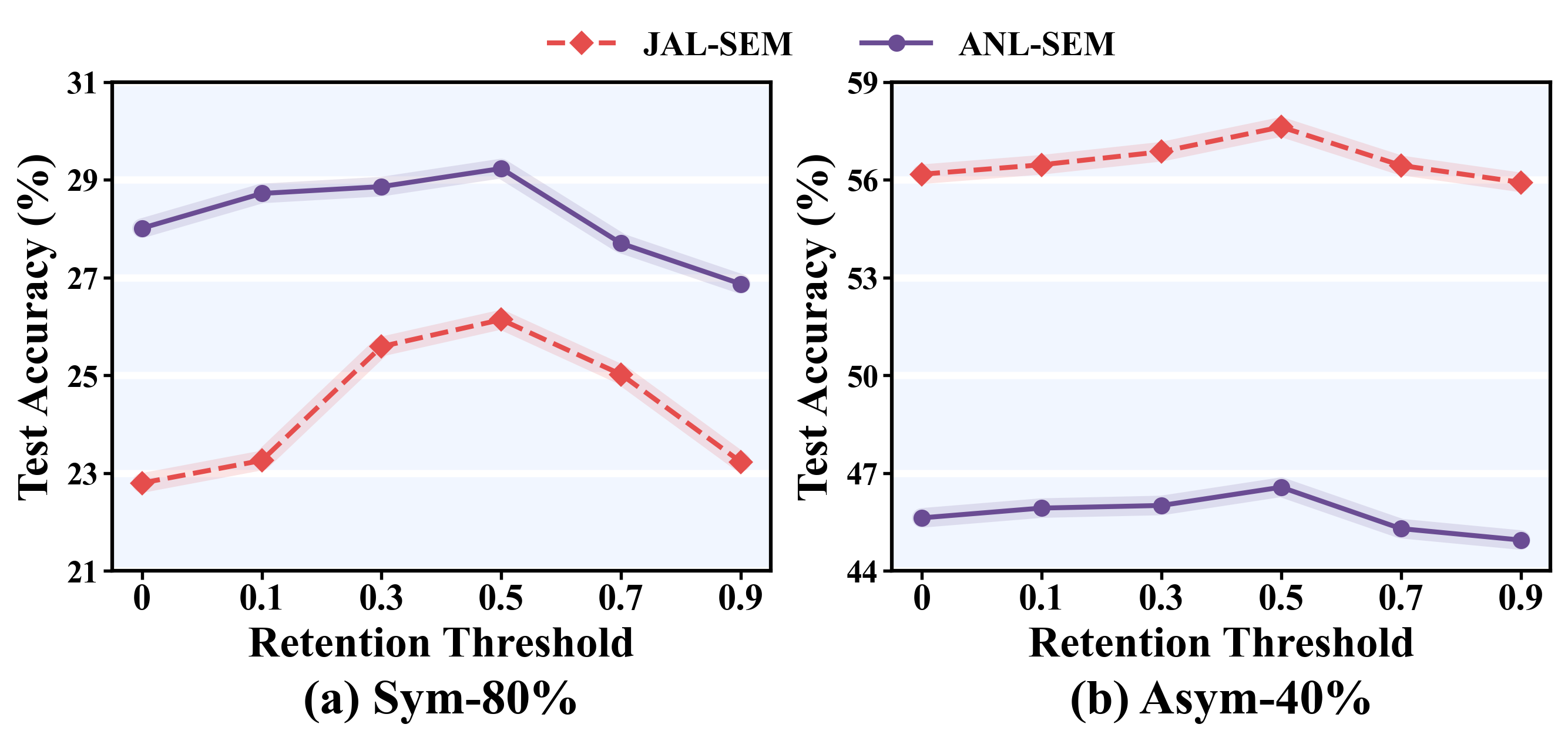}
    \vspace{-0.6cm}
    \caption{ 
    Sensitivity analysis of different retention thresholds on CIFAR-100. 
    The average test accuracies (\%) are reported over 3 random runs.
    } 
    \label{fig5}
    \vspace{-0.2cm}
\end{figure}

\textbf{Retention Threshold Analysis.} 
We investigate the sensitivity of SEM to the retention threshold $\rho$ on CIFAR-100. The hyperparameter $\rho$ determines which edges are retained deterministically, with a larger value inducing more aggressive masking. We use $\rho=0$ as the baseline, corresponding to the standard JAL and ANL methods without SEM. As illustrated in Fig.~\ref{fig5}, performance generally increases and then decreases as $\rho$ increases across different methods and noise conditions. Specifically, a small $\rho$ fails to mask sufficient redundant connections, limiting the model\textquotesingle s ability to suppress the backpropagation of noisy gradients. Conversely, a large $\rho$ may discard critical edges and impair the model\textquotesingle s capacity to fit clean data. Accordingly, we fix $\rho=0.5$ across all experiments without dataset-specific retuning.

\textbf{Applicability to KAN-Based Classifiers.}
To examine the applicability of SEM beyond conventional FC classifiers, we extend it to a KAN classifier and denote the resulting SEM-enhanced layer as SKAN. For sample $i$, the contribution of input feature $k$ to class $j$ is defined as $a_{ijk}=\phi_{jk}(v_{ik})$, where $\phi_{jk}$ is the corresponding learnable univariate function. During training, we apply the RMS scoring and \textit{Bernoulli} masking rules in Eqs.~(13)--(16) to these edge contributions. CE-KAN and CE-SKAN employ KAN and SKAN classifiers, respectively. As shown in Table~\ref{tab9}, CE-SKAN outperforms CE-KAN across the evaluated noise conditions, supporting the applicability of SEM beyond conventional FC classifiers.

\section{Conclusion}
\label{sec6}
In this study, we draw on Optimal Brain Damage (OBD) theory to identify low-activation classifier connections whose masking induces limited estimated loss perturbation. We further show that classifier masking tightens a derived norm-based upper bound on noise-induced gradient error. Guided by this principle, we propose a novel selective edge masking (SEM) mechanism from a simple classifier-level architectural adaptation perspective for noisy label learning. Through a systematic experimental analysis of model robustness, we reveal that our SEM can effectively mitigate gradient errors propagated from noisy labels while simultaneously preserving the model\textquotesingle s capacity to fit clean samples. Additionally, the improvements achieved by applying SEM to KAN-based classifiers support the applicability of SEM beyond conventional FC classifiers. Comprehensive experiments integrating SEM with various noise-robust methods demonstrate its effectiveness across the evaluated noisy label learning settings. In future work, we plan to extend SEM to more sophisticated architectures, such as Transformers, and adapt it to other noise-robust learning tasks, including noisy correspondence.

\begin{table}[!t]
\centering
\setlength{\tabcolsep}{8.5pt} % Adjust column spacing
\renewcommand{\arraystretch}{0.97}
\caption{
Effect of each component for SEM under Sym-80\%. \textit{Bernoulli} (0.5) indicates that edges are retained with a fixed probability of 0.5, while \textit{Bernoulli} ($\bar{\textbf{\textit{S}}}$) represents our importance-based sampling strategy.
}
\vspace{-0.1cm}
\begin{tabular}{lccc}
\toprule
Method & Component & CIFAR-10 & CIFAR-100 \\ 
\midrule
CE & baseline & 19.74±0.40 & 7.82±0.33\\ 
\multirow{4}{*}{CE-SEM} 
& w/o EMA  & 39.25±0.86 & 17.59±0.31\\
& w/o \textit{Bernoulli} & 21.98±0.79 & 7.96±0.89 \\ 
& with \textit{Bernoulli} (0.5) & 23.53±0.57 & 8.09±0.65\\
& \cellcolor[HTML]{DAE8FC}\textbf{with \textit{Bernoulli} ($\bar{\textbf{\textit{{S}}}}$)} & \cellcolor[HTML]{DAE8FC}\textbf{39.72±0.81}  & \cellcolor[HTML]{DAE8FC}\textbf{18.63±0.29} \\
\bottomrule
\end{tabular}
\label{tab8}
\vspace{-0.4cm}
\end{table}

\begin{table}[!t]
\centering
\caption{
Applicability to KAN-based classifiers. CE-KAN uses a KAN layer as the classifier, whereas CE-SKAN uses its SEM-enhanced variant.
}
\vspace{-0.2cm}
\begin{tabular}{l|cc|cc}
\toprule
\multirow{2}{*}{\textbf{Methods}} & \multicolumn{2}{c|}{\textbf{CIFAR-10}} & \multicolumn{2}{c}{\textbf{CIFAR-100}} \\
\cmidrule{2-5}
& Sym-80\% & Asym-40\% & Sym-80\% & Asym-40\% \\
\midrule
CE-KAN & 17.71±1.33 & 73.68±0.63  & 8.18±0.54 & 40.95±0.73\\ 
\cellcolor[HTML]{DAE8FC}\textbf{CE-SKAN} & \cellcolor[HTML]{DAE8FC}\textbf{38.54±1.56} & \cellcolor[HTML]{DAE8FC}\textbf{80.03±0.75} & \cellcolor[HTML]{DAE8FC}\textbf{12.13±0.19} & \cellcolor[HTML]{DAE8FC}\textbf{43.04±0.49}\\
\bottomrule
\end{tabular}
\label{tab9}
\vspace{-0.4cm}
\end{table}

\section*{Acknowledgments}
This work was supported in part by the National Natural Science Foundation of China under Grant 62302149 and 62372155, in part by Basic Research Program of Jiangsu under Grant BK20250188, in part by the Research Funds of Jiangsu Hydraulic Research Institute under Grant 2025z065,  in part by the Major Science and Technology Program of the Ministry of Water Resources of China under Grant SKS-2022072 and in part by the China Postdoctoral Science Foundation under Grant 2025M771578.

\bibliographystyle{IEEEtran}
\bibliography{ref}

\clearpage
{\appendices

\setcounter{figure}{0}
\setcounter{table}{0}
\setcounter{equation}{0}

\renewcommand{\thefigure}{S\arabic{figure}}
\renewcommand{\thetable}{S\arabic{table}}
\renewcommand{\theequation}{S\arabic{equation}}

\section*{Overview}

\begin{figure}[!h]
    \centering
    \includegraphics[width=0.98\linewidth]{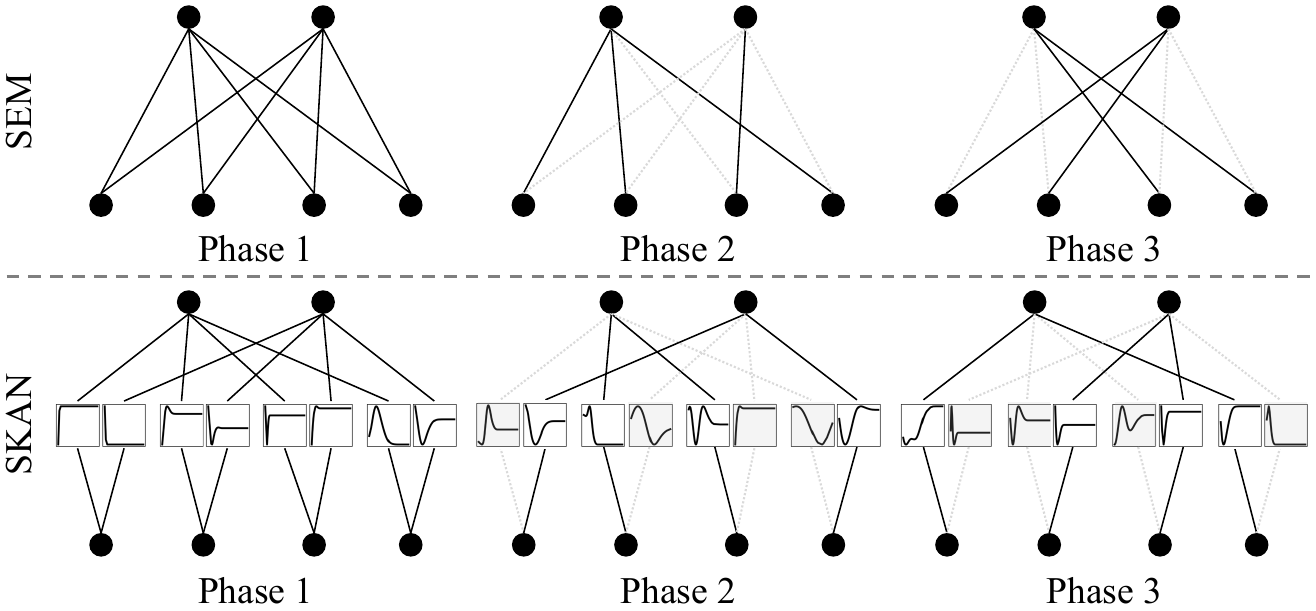}
    \caption{Visualization of SEM and SKAN (extending SEM to KAN). During training, connection patterns dynamically adapt to optimize information flow, thereby propagating critical features and improving noise robustness.} 
    \label{figs1}
\end{figure}

This supplement provides implementation details, additional quantitative and qualitative results, and the SEM algorithm.

\section*{Implementation Details}
\textbf{Noise generation.} 
We generate synthetic labels for CIFAR-10 and CIFAR-100. For symmetric noise, labels within each class are randomly flipped to other classes. For asymmetric noise, labels are flipped within semantically related classes. In CIFAR-10, the flips are: $\text{BIRD} \rightarrow \text{AIRPLANE}$, $\text{CAT} \rightarrow \text{DOG}$, $\text{DEER} \rightarrow \text{HORSE}$, and $\text{TRUCK} \rightarrow \text{AUTOMOBILE}$. In CIFAR-100, classes are grouped into 20 super-classes (each containing 5 sub-classes), and labels are circularly flipped to the next adjacent class within the same super-class. 
For instance-dependent noise, we follow the approach in PDN.

\textbf{Training configurations.} Table~\ref{tabs6} details the training settings, with SEM aligned to each baseline\textquotesingle s configuration.
% To ensure a fair comparison, we strictly align the training parameters of SEM with other baselines according to their respective original papers. The detailed training settings are provided in Table~\ref{tabs6}.

\section*{Quantitative Results}
\textbf{Suppression Scope Analysis.} 
Since our edge-wise masking is not directly applicable to CNN layers, we use CDR to assess whether full-network suppression is necessary. The original CDR suppresses updates to non-critical parameters across the full network, whereas our modified variant restricts it to the FC classifier. Table~\ref{tabs1} demonstrates that restricting suppression to the FC classifier achieves comparable or better performance, supporting its sufficiency for noisy label learning.

\begin{figure}[!t]
    \centering
    \includegraphics[width=0.95\linewidth]{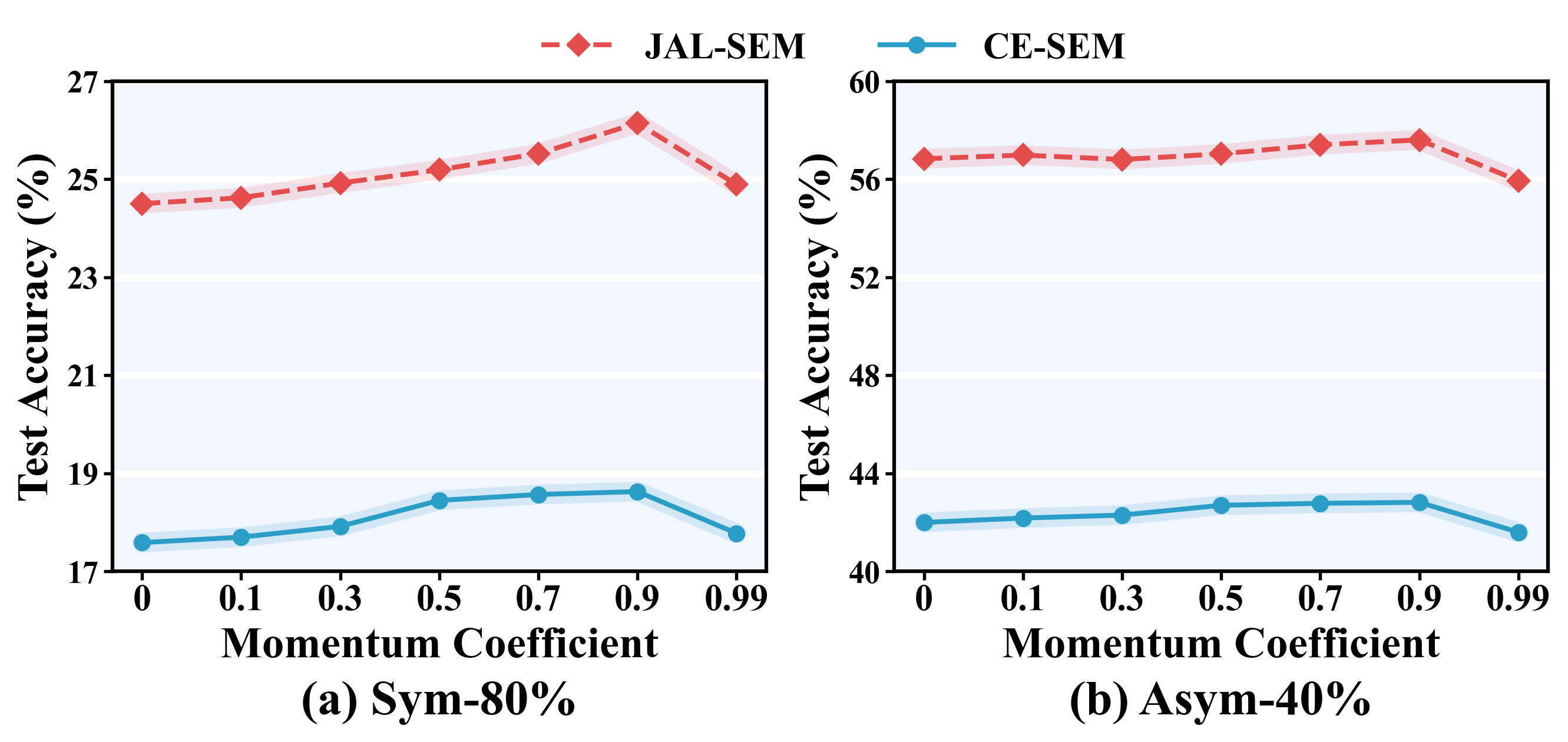}
    \vspace{-0.2cm}
    \caption{ 
    Sensitivity analysis of different momentum coefficient $\beta$ on CIFAR-100. 
    The average test accuracies (\%) are reported over 3 random runs.
    } 
    \label{figs2}
    \vspace{-0.2cm}
\end{figure}

\textbf{Momentum Coefficient Sensitivity.}
We evaluate SEM sensitivity to $\beta$ on CIFAR-100. Since $\beta$ controls EMA smoothing, a larger value yields smoother importance estimations. Fig.~\ref{figs2} indicates that SEM is robust across $\beta \in (0, 0.9]$. However, a larger $\beta$ (\ie, $0.99$) degrades performance by impeding adaptation to current learning dynamics. Thus, we set $\beta=0.9$.
% We evaluate SEM sensitivity to different $\beta$ values on CIFAR-100. Since $\beta$ controls the EMA smoothing degree, a larger value yields smoother edge importance estimations. As shown in Fig.~\ref{figs2}, SEM exhibits robustness across $\beta \in (0, 0.9]$. However, a larger $\beta$ (\ie, $0.99$) degrades performance by impeding model adaptation to current learning dynamics. Thus, we set $\beta=0.9$.

\textbf{Deterministic Retention Strategy Analysis.} 
Table~\ref{tabs2} shows that removing the deterministic retention mechanism (w/o $\rho$) degrades performance. The result indicates that importance-based \textit{Bernoulli} masking alone risks discarding crucial pathways. Instead, retaining edges with $\bar{s}_{jk}^{(t)} \ge \rho$ preserves discriminative capacity, thereby stabilizing training and enhancing noise robustness.
% Table~\ref{tabs2} shows that removing the deterministic retention mechanism (w/o $\rho$) yields suboptimal performance relative to full SEM. The result indicates that importance-based \textit{Bernoulli} masking alone risks discarding crucial pathways. Instead, retaining the most critical edges (\ie, $\bar{s}_{jk}^{(t)} \ge \rho$) provides a necessary structural constraint to preserve discriminative capacity, thereby stabilizing training and enhancing noise robustness.
% As presented in Table~\ref{tabs2}, removing the deterministic retention mechanism (w/o $\rho$) yields suboptimal performance compared to the full SEM. This result indicates that relying solely on importance-based \textit{Bernoulli} masking risks discarding crucial information pathways for training. In stead, preserving the most critical edges (\ie, $\bar{s}_{jk}^{(t)} \ge \rho$) acts as a necessary structural constraint to maintain the network\textquotesingle s discriminative capacity, thereby stabilizing training and enhancing noise robustness.

\textbf{Batch Size Sensitivity.} 
Since SEM estimates importance across mini-batches, we evaluate its sensitivity to batch size on CIFAR-10. Fig.~\ref{figs3} shows that the EMA strategy (\ie, $\beta=0.9$) effectively stabilizes scores across mini-batches, yielding minor variations across batch sizes. Furthermore, a larger batch size (\ie, 256) slightly improves performance via more stable importance estimates for masking. Overall, SEM is robust to batch size. For fairness, we align it with all baseline methods.
% Since SEM relies on cross-sample importance estimation, we evaluate its sensitivity to batch size on CIFAR-10. As illustrated in Figure~\ref{figs3}, the EMA strategy (\ie, $\beta=0.9$) effectively stabilizes importance scores across mini-batches, ensuring minor variations across different batch sizes. Furthermore, a larger batch size (\ie, 256) results in a slight performance gain by producing a more stable importance estimate for masking. The result indicates that our SEM is robust to batch size. For a fair comparison, we align it with all baselines throughout our experiments.

\begin{table}[!t]
\centering
\setlength{\tabcolsep}{8pt} % Adjust column spacing
\caption{Comparison of CDR performance under different suppression scopes. "Full Network" suppresses gradient updates for all unimportant parameters across the entire network, while "FC Classifier" masks only non-critical edges in the last FC layer.}
\begin{tabular}{lccc}
\toprule
Datasets & Suppression Scope & Sym-80\% & Asym-40\%\\
\midrule
\multirow{2}{*}{CIFAR-10} 
& Full Network & 33.73±0.60 & 75.59±0.46 \\
& FC Classifier & 33.61±0.48 & 76.05±0.47\\
\midrule
\multirow{2}{*}{CIFAR-100}
& Full Network & 16.04±0.19 & 40.88±0.69 \\
& FC Classifier & 16.94±0.75 & 41.18±0.41 \\
\bottomrule
\end{tabular}
\label{tabs1}
\vspace{-0.4cm}
\end{table}

\textbf{Matched-Sparsity Comparison.}
To assess whether SEM\textquotesingle s gains stem merely from reduced connectivity, we compare SEM-TopK with Dropout and DropConnect at a target retention rate of 50\% under identical training settings. Dropout and DropConnect retain neurons and edges with 50\% probability, respectively, whereas SEM-TopK deterministically retains the top 50\% classifier edges ranked by the EMA-smoothed scores $\bar{\boldsymbol{S}}^{(t)}$. As shown in Table~\ref{tabs3}, SEM-TopK consistently outperforms both baselines, supporting the effectiveness of EMA-guided edge selection beyond random sparsification.

\textbf{Inference Scaling Analysis.} 
Stochastic regularizers (\eg, Dropout) typically employ weight scaling to maintain expected outputs. Since SEM dynamically masks edges via EMA-smoothed score $\bar{\boldsymbol{S}}^{(t)}$, we introduce a scaling matrix $\boldsymbol{E}^{(t)} = \{e_{jk}^{(t)}\}$ to match the expected output for  inference:
\begin{equation}
e_{jk}^{(t)} = \begin{cases} 1, & \bar{s}_{jk}^{(t)} \geq \rho, \\ \bar{s}_{jk}^{(t)}, & 0 \leq \bar{s}_{jk}^{(t)} < \rho. \end{cases}
\end{equation}
Table~\ref{tabs4} shows that applying expectation-matched inference scaling (\ie, $\boldsymbol{W} \odot \boldsymbol{E}^{(t)}$) unexpectedly degrades performance. This counterintuitive result aligns with our theoretical analysis (Proposition 1). Since SEM primarily masks redundant, low-activation edges, it minimally influences the final logits and implicitly concentrates optimization on retained critical pathways during training. 
Thus, imposing scaling could suppress feature information and disrupt the learned representation, confirming that SEM requires no explicit inference scaling.

\begin{figure}[!t]
    \centering
    \includegraphics[width=0.95\linewidth]{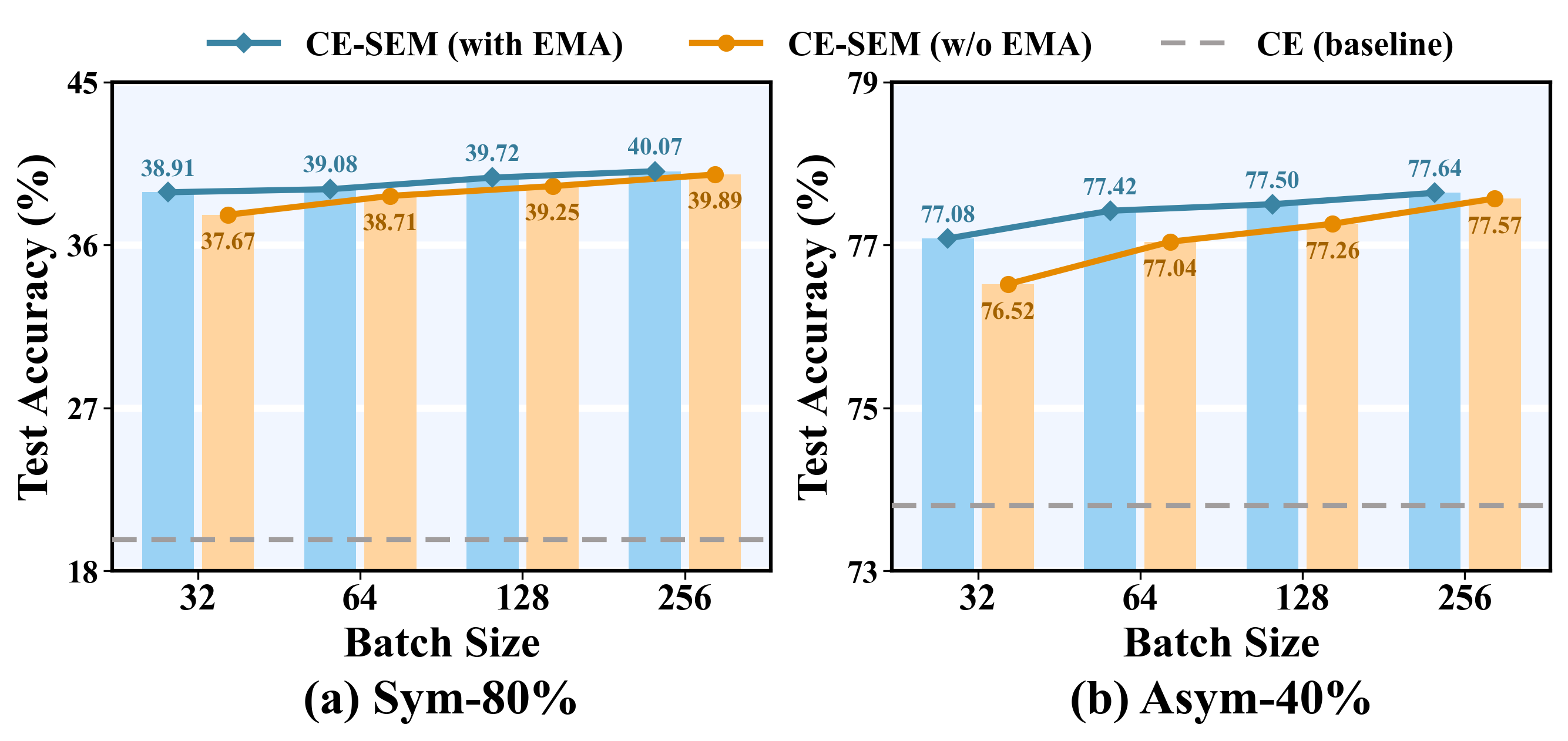}
    \vspace{-0.3cm}
    \caption{Sensitivity analysis of different batch sizes on CIFAR-10. 
    The average test accuracies (\%) are reported over 3 random runs.
    } 
    \label{figs3}
    \vspace{-0.2cm}
\end{figure}

\begin{table}[!t]
\setlength{\tabcolsep}{11pt} % Adjust column spacing
\centering
\caption{Effect of deterministic retention strategy with the threshold $\rho$ under Sym-80\%. The average test accuracies (\%) are reported over 3 random runs.
% Comparison between the standard CE-SEM (with $\rho$) and a variant relying solely on the importance-based Bernoulli sampling strategy (w/o $\rho$).
}
\vspace{-0.2cm}
\begin{tabular}{lccc}
\toprule
Method & Component & CIFAR-10 & CIFAR-100 \\ 
\midrule
CE & baseline & 19.74±0.40 &  7.82±0.33 \\ 
% \midrule
\multirow{2}{*}{CE-SEM} & w/o $\rho$ & 39.04±1.36 & 17.97±0.35\\ 
 & \cellcolor[HTML]{DAE8FC} \textbf{with $\rho$} & \cellcolor[HTML]{DAE8FC}\textbf{39.72±0.81} & \cellcolor[HTML]{DAE8FC}\textbf{18.63±0.29}\\ 
 \midrule
 JAL & baseline & 65.43±0.99 & 22.80±2.11\\
 \multirow{2}{*}{JAL-SEM} & w/o $\rho$ & 64.30±0.61 & 24.62±2.03\\ 
 & \cellcolor[HTML]{DAE8FC} \textbf{with $\rho$} & \cellcolor[HTML]{DAE8FC}\textbf{66.39±1.08} & \cellcolor[HTML]{DAE8FC}\textbf{26.14±2.77}\\ 
\bottomrule
\end{tabular}
\label{tabs2}
\vspace{-0.2cm}
\end{table}

\begin{table}[!h]
\setlength{\tabcolsep}{5.5pt}
\centering
\caption{Matched-sparsity comparison under Sym-80\% noise with a target classifier-edge retention ratio of 50\%. The average test accuracies (\%) are reported over 3 random runs.}
\vspace{-0.2cm}
\begin{tabular}{lccc}
\toprule
Method & Masking Rule & CIFAR-10 & CIFAR-100 \\
\midrule
CE-Dropout
& Random Neuron & 29.96±1.09 & 7.29±0.37 \\
CE-DropConnect
& Random Edge & 23.09±0.25 & 7.34±0.16 \\
\textbf{CE-SEM-TopK}
& \textbf{EMA-ranked Edge} & \textbf{31.34±0.51} & \textbf{12.91±0.56} \\
\bottomrule
\end{tabular}
\label{tabs3}
\vspace{-0.2cm}
\end{table}

\textbf{Computational Overhead Analysis.} 
Table~\ref{tabs5} reports trainable parameters, floating-point operations (FLOPs), and per-sample inference latency. SEM introduces no additional trainable parameters and maintains nearly identical inference FLOPs, while edge scoring and masking are confined to training. Although these operations are disabled at inference, the customized classifier implementation used in our experiments introduces a minor latency increase from 1.95 ms to 2.16 ms. Overall, SEM incurs minor inference overhead.
% Table~\ref{tabs5} reports trainable parameters, floating-point operations (FLOPs), and per-sample inference latency. 
% Our SEM introduces no trainable parameters and nearly identical inference FLOPs. Although our SEM is disabled at inference, the customized classifier forward pass slightly increases latency from 1.95 ms to 2.16 ms. This minimal overhead is far outweighed by the notable performance gains of our SEM for noisy label learning.
% Our ResNet34-SEM maintains nearly identical parameter counts and FLOPs. While our inference latency slightly increases (from 1.95 ms to 2.16 ms) due to element-wise masking in the forward pass, this minimal overhead is far outweighed by the notable performance gains of our SEM for noisy label learning.

\section*{Qualitative Results}
\textbf{Selective Edge Masking Process.} 
Fig.~\ref{figs1} visualizes our selective edge masking process in FC and SKAN classifiers. Starting from full connectivity, our mechanism induces high sparsity by temporarily discarding low-importance edges at each iteration. This adaptive process enables the model to focus on critical connections and enhance noise robustness.
% Fig.~\ref{figs4} visualizes our selective edge masking process for both FC and SKAN classifiers. Beginning with a fully connected model, training with SEM renders it highly sparse. Specifically, edges with low importance scores are temporarily discarded during each iteration. This adaptive masking mechanism enables the model to focus on critical connections and enhance noise robustness.

\textbf{Adaptive Edge Retention Ratio. }
Fig.~\ref{figs4} tracks the classifier edge retention ratio during CIFAR-100 training. Unlike Dropout or DropConnect, SEM dynamically reduces this ratio until it stabilizes at a low level. This trend shows that SEM discards most non-critical connections late in training, restricting parameter updates and mitigating noise memorization.
% Fig.~\ref{figs4} visualizes the classifier edge retention ratio during CIFAR-100 training. Unlike Dropout or DropConnect, SEM dynamically adjusts this ratio, which gradually decreases and stabilizes at a low level. This observation indicates that SEM effectively discards most non-critical connections during late training, thereby restricting parameter updates to mitigate noise memorization.

\textbf{CAM Visualization. }
Figures~\ref{figs5} and \ref{figs6} present CAM visualizations on WebVision-Mini and ILSVRC12. Compared to standard CE, CE-SEM highlights more precise regions of interest, visually supporting its enhanced noise robustness.

\begin{figure}[!t]
    \centering
    \includegraphics[width=0.95\linewidth]{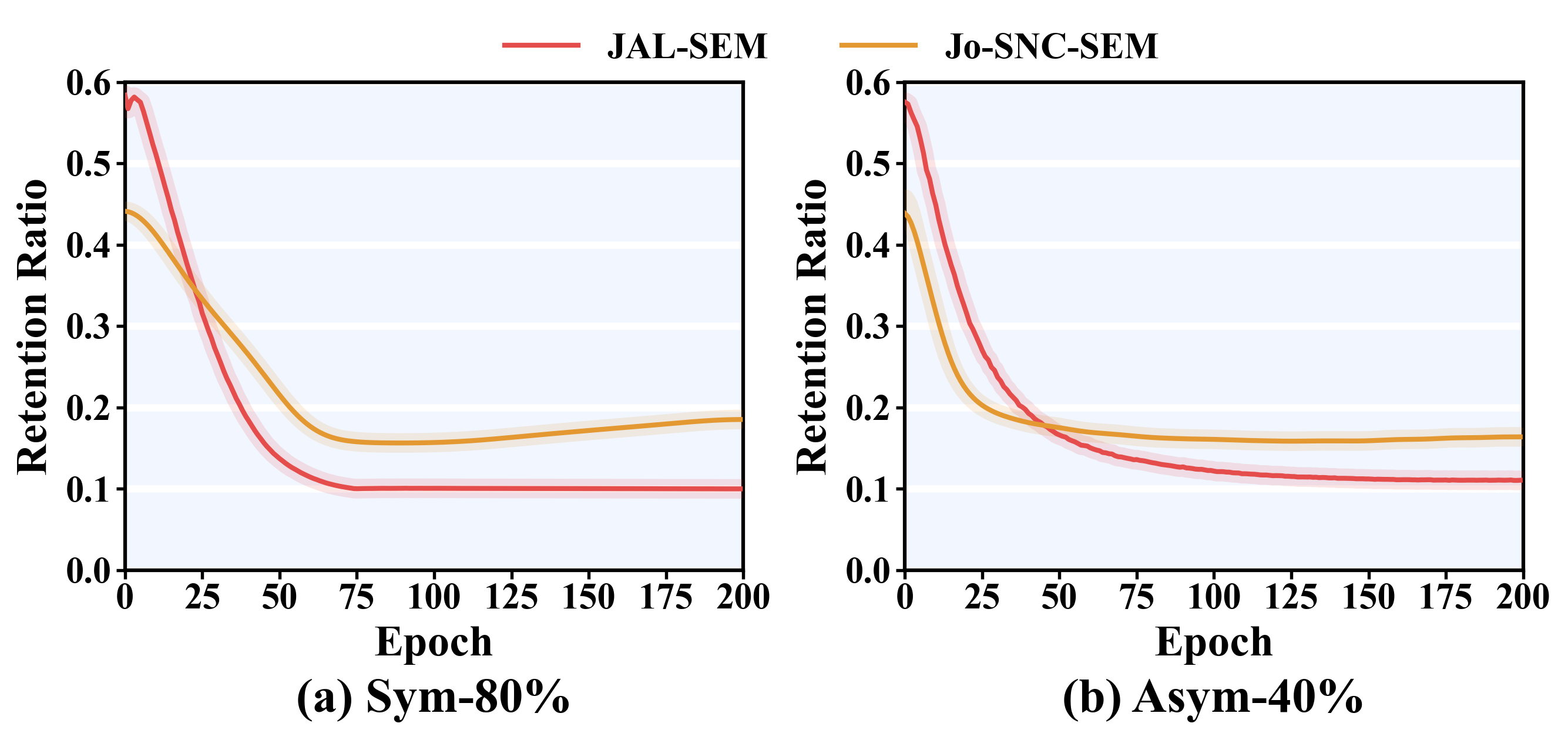}
    \vspace{-0.3cm}
    \caption{ 
    Adaptive edge retention ratio on CIFAR-100. Instead of a fixed ratio, our SEM adaptively adjusts the retention ratio during training. 
    } 
    \label{figs4}
    \vspace{-0.2cm}
\end{figure}

\begin{table}[!h]
\setlength{\tabcolsep}{7.8pt} % Adjust column spacing
\centering
\caption{
Effect of inference scaling on CIFAR-10 and CIFAR-100 under Sym-80\%. The average test accuracies (\%) are reported over 3 random runs.
% The test accuracy over 3 random runs is reported. 
% The blue-highlighted region represents the best approach.
}
\vspace{-0.2cm}
\begin{tabular}{lccc}
\toprule
Method & Inference Scaling & CIFAR-10 & CIFAR-100 \\ 
\midrule
CE & \multicolumn{1}{c}{baseline} & 19.74±0.40 & 7.82±0.33 \\ 
% \midrule
\multirow{2}{*}{CE-SEM}
 & $\checkmark$ &  37.25±0.69 & 15.94±0.76\\ 
 & \cellcolor[HTML]{DAE8FC}\textbf{$\times$} & \cellcolor[HTML]{DAE8FC}\textbf{39.72±0.81} & \cellcolor[HTML]{DAE8FC}\textbf{18.63±0.29}\\
 \midrule
 JAL & baseline & 65.43±0.99 & 22.80±2.11\\
 \multirow{2}{*}{JAL-SEM} & $\checkmark$ & 65.71±1.03 & 25.47±2.54\\ 
 & \cellcolor[HTML]{DAE8FC}\textbf{$\times$} & \cellcolor[HTML]{DAE8FC}\textbf{66.39±1.08} & \cellcolor[HTML]{DAE8FC}\textbf{26.14±2.77}\\ 
\bottomrule
\end{tabular}
\label{tabs4}
\vspace{-0.2cm}
\end{table}

\begin{table}[!h]
\centering
\caption{Comparison of parameter counts, FLOPs, and inference latency across different models.}
\vspace{-0.2cm}
\begin{tabular}{cccc}
\toprule
Backbone & Parameters (M) & FLOPs (G) & Latency (ms) \\
\midrule
ResNet34 & 21.33 & 0.5818 & 1.95\\
% \midrule
ResNet34-SEM & 21.33 & 0.5819 & 2.16\\
\bottomrule
\end{tabular}
\label{tabs5}
\vspace{-0.4cm}
\end{table}

\section*{Algorithm Details} 
Algorithm~\ref{algorithm1} details our SEM. During training, it adaptively masks non-critical edges to suppress noisy gradient propagation. During inference, the masking operation is disabled, allowing the model to leverage its full capacity for prediction.

\clearpage
\begin{figure*}[!t]
    \centering
    \includegraphics[width=1.0\linewidth]{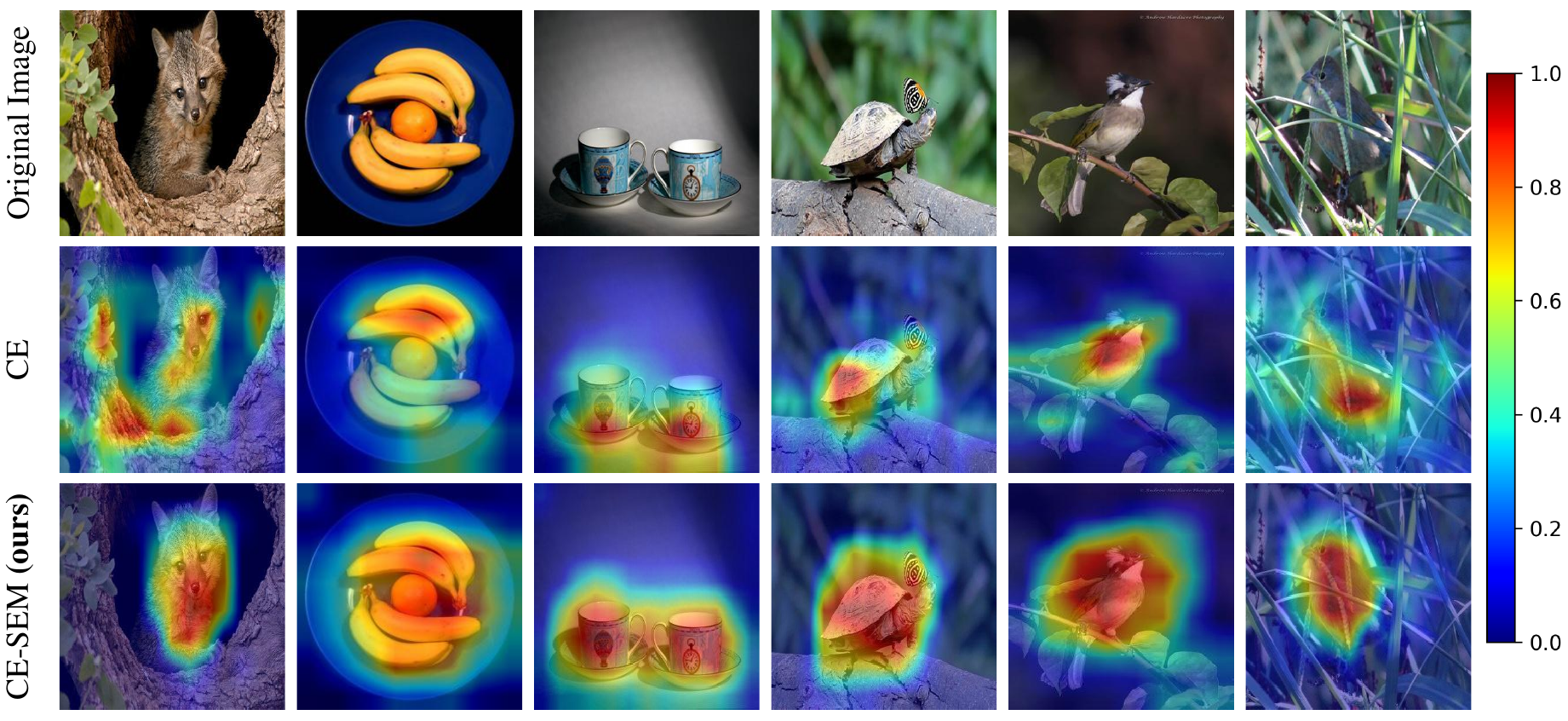}
    % \vspace{-0.6cm}
    \caption{ 
    Class activation map (CAM) visualizations of different models trained on the WebVision-Mini dataset. The samples are drawn from the WebVision-Mini validation set. In the CAMs, color represents the intensity of feature activation, where red indicates high activation levels (focus regions) and blue denotes non-activated background areas.
    } 
    \label{figs5}
\end{figure*}

\begin{figure*}[!t]
    \centering
    \includegraphics[width=1.0\linewidth]{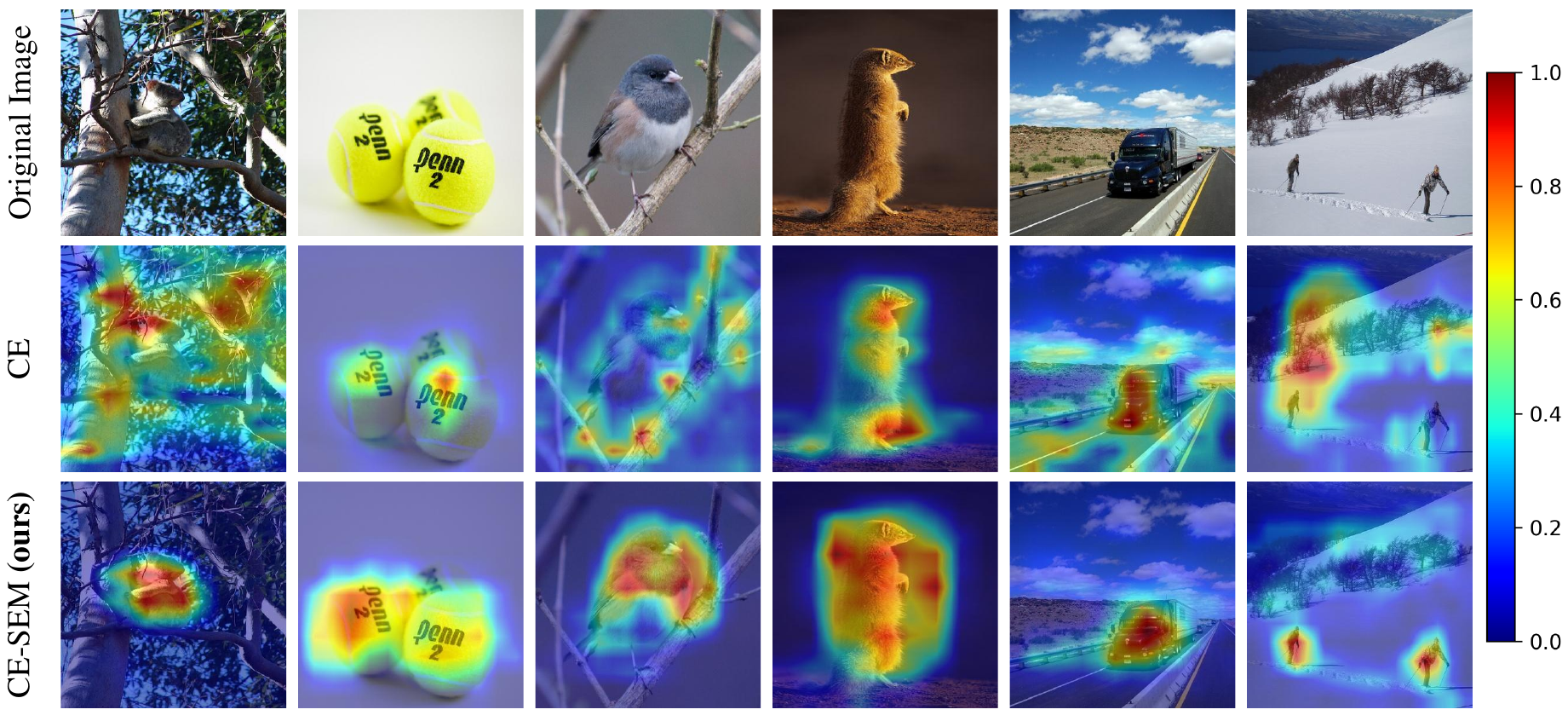}
    % \vspace{-0.6cm}
    \caption{ 
    Class activation map (CAM) visualizations of different models trained on the WebVision-Mini dataset. The samples are drawn from the ILSVRC12 validation set. In the CAMs, color represents the intensity of feature activation, where red indicates high activation levels (focus regions) and blue denotes non-activated background areas.
    } 
    \label{figs6}
\end{figure*}

\clearpage

\begin{table*}[t]
\centering
\caption{Training parameters across various datasets with different methods. The "Cosine" denotes the Cosine Annealing scheduler, and “pt” indicates a pre-trained model. StepLR ($s$,$\gamma$) multiplies the current learning rate by $\gamma$ every $s$ epochs, while "\(e\)-th" indicates a $0.1\times lr$ decay at epoch $e$.
% The notation "pt" indicates the utilization of a pre-trained model. 
When integrating with different approaches, our SEM follows their original training configurations. 
}
\setlength{\tabcolsep}{2pt}
\begin{tabular}{c|c|cc|cc|cc}
\toprule
\multirow{2}{*}{Params} & CIFAR-10 & \multicolumn{2}{c|}{CIFAR-100 \& CIFAR80N-O} & \multicolumn{2}{c|}{WebVision-Mini} & \multicolumn{2}{c}{Clothing1M} \\
\cmidrule{2-8}
&  CE \& ANL \& JAL & CE \& ANL \& JAL & Jo-SNC & JAL & Jo-SNC & JAL &  DivideMix \\
\cmidrule{1-8}
model & 8-layer CNN & ResNet-34 & 7-layer CNN &  InceptionResNetV2 & InceptionResNetV2 & ResNet-50(pt) & ResNet-50(pt) \\ 
epochs & 120 & 200 & 200 & 250 & 100 & 10 & 80 \\
batch size & 128 & 128 & 128 & 256 & 64 & 256 & 32 \\
lr & 0.01 & 0.1 & 0.001 & 0.4 & 0.01 & 0.005 & 0.002 \\
scheduler &  Cosine & Cosine &  Cosine & StepLR (1, 0.97) & 20th \& 40th \& 60th \& 80th & StepLR (5, 0.1) & 40th \\
wd & 1e-4 & 1e-5 & 0 & 3e-5 & 1e-4 & 1e-3 & 1e-4\\
\bottomrule
\end{tabular}
\label{tabs6}
\end{table*}

\definecolor{mydeepblue}{RGB}{0,0,139} % RGB 深蓝色 

\definecolor{mydeepblue}{RGB}{0,0,139} % 

\begin{algorithm*}[t]
\caption{Selective Edge Masking (SEM)}
\label{algorithm1}
\begin{algorithmic}[1]

\Require
$\widetilde{\mathcal{D}}_{\mathrm{train}}$: Noisy training dataset;
$\mathcal{D}_{\mathrm{test}}$: Testing dataset;
$\psi$: Visual backbone network;
$\bm{W}$: Learnable weight matrix of the FC classifier;
$\bm{b}$: Learnable bias vector of the FC classifier;
$\rho$: Retention threshold;
$\beta$: Momentum coefficient.

\Statex
\textit{\textcolor{mydeepblue}{\# Training Phase}}

\For{each training epoch}
    \For{each mini-batch
    $(\bm{X}_\mathrm{train},\widetilde{\bm{Y}})
    \subset\widetilde{\mathcal{D}}_{\mathrm{train}}$}

        \Statex
        \hspace{\algorithmicindent}
        \hspace{\algorithmicindent}
        \textit{\textcolor{mydeepblue}
        {\# Stage 1: Edge Importance Scoring}}

        \State Extract the input features:
        $\bm{V}\gets\psi(\bm{X}_\mathrm{train})$;

        \State Calculate the edge activation tensor
        $\bm{A}=[a_{ijk}]$, where
        $a_{ijk}=v_{ik}w_{jk}$;

        \State Compute the edge importance score via Root Mean Square:
        $\bm{S}\gets\operatorname{RMS}(\bm{A})$;
        \Comment{\textcolor{mydeepblue}{Eq.~(13)}}

        \State Compute the normalized edge importance score:
        $\widehat{\bm{S}}
        \gets\operatorname{Normalization}(\bm{S})$;
        \Comment{\textcolor{mydeepblue}{Eq.~(14)}}

        \State Compute the EMA-smoothed importance score:
        $\bar{\bm{S}}\gets
        \operatorname{EMA}
        (\bar{\bm{S}},\widehat{\bm{S}},\beta)$,
        with $\bar{\bm{S}}^{(0)}=\widehat{\bm{S}}^{(0)}$;
        \Comment{\textcolor{mydeepblue}{Eq.~(15)}}

        \Statex
        \hspace{\algorithmicindent}
        \hspace{\algorithmicindent}
        \textit{\textcolor{mydeepblue}
        {\# Stage 2: Edge Masking}}

        \State Generate the binary mask matrix:
        $\bm{M}\gets
        \mathbb{I}(\bar{\bm{S}}\geq\rho)
        +
        \mathbb{I}(\bar{\bm{S}}<\rho)
        \odot
        \operatorname{Bernoulli}(\bar{\bm{S}})$;
        \Comment{\textcolor{mydeepblue}{Eq.~(16)}}

        \State Obtain the masked weight matrix:
        $\bar{\bm{W}}\gets\bm{M}\odot\bm{W}$;

        \State Compute the output logits:
        $\bm{Z}\gets
        \bm{V}\bar{\bm{W}}^{\mathrm{T}}+\bm{b}$;

        \State Compute the loss
        $\mathcal{L}(\bm{Z},\widetilde{\bm{Y}})$
        and update the model parameters via backpropagation.

    \EndFor
\EndFor

\Statex
\textit{\textcolor{mydeepblue}{\# Inference Phase}}

\For{each mini-batch
$(\bm{X}_{\mathrm{test}},\bm{Y}_{\mathrm{test}})
\subset\mathcal{D}_{\mathrm{test}}$}

    \State Extract the input features:
    $\bm{V}\gets\psi(\bm{X}_{\mathrm{test}})$;

    \State Restore full connectivity and compute the output logits:
    $\bm{Z}\gets
    \bm{V}\bm{W}^{\mathrm{T}}+\bm{b}$;

    \State Compute and store the predictions:
    $\widehat{\bm{Y}}_{\mathrm{test}}
    \gets\operatorname{argmax}(\bm{Z})$;

\EndFor

\State Compute the test accuracy using
$\widehat{\bm{Y}}_{\mathrm{test}}$
and $\bm{Y}_{\mathrm{test}}$.

\end{algorithmic}
\end{algorithm*}

}

\end{document}